\documentclass{article}

\usepackage{microtype}
\usepackage{graphicx}
\usepackage{subfig}
\usepackage{booktabs} %

\usepackage{multirow}
\usepackage[svgnames]{xcolor}

\usepackage{hyperref}

\usepackage[accepted]{icml2025}

\usepackage{amsmath}
\usepackage{amssymb}
\usepackage{mathtools}
\usepackage{amsthm}

\usepackage[capitalize,noabbrev]{cleveref}

\theoremstyle{plain}
\newtheorem{theorem}{Theorem}[section]
\newtheorem{proposition}[theorem]{Proposition}

\theoremstyle{definition}

\theoremstyle{remark}

\usepackage[textsize=tiny]{todonotes}

\icmltitlerunning{HyperbolicLR: Epoch Insensitive Learning Rate Scheduler}

\begin{document}

\twocolumn[
\icmltitle{HyperbolicLR: Epoch Insensitive Learning Rate Scheduler}

\begin{icmlauthorlist}
\icmlauthor{Tae-Geun Kim}{yyy}
\end{icmlauthorlist}

\icmlaffiliation{yyy}{Department of Physics, Yonsei University, Seoul, Republic of Korea}
\icmlcorrespondingauthor{Tae-Geun Kim}{tg.kim@yonsei.ac.kr}

\icmlkeywords{Learning Rate Scheduling, Deep Learning, Optimization, Neural Networks, Hyperbolic Functions, ICML}

\vskip 0.3in
]

\printAffiliationsAndNotice{\icmlEqualContribution} %

\begin{abstract}
This study proposes two novel learning rate schedulers---Hyperbolic Learning Rate Scheduler (HyperbolicLR) and Exponential Hyperbolic Learning Rate Scheduler (ExpHyperbolicLR)---to address the epoch sensitivity problem that often causes inconsistent learning curves in conventional methods. By leveraging the asymptotic behavior of hyperbolic curves, the proposed schedulers maintain more stable learning curves across varying epoch settings. Specifically, HyperbolicLR applies this property directly in the epoch-learning rate space, while ExpHyperbolicLR extends it to an exponential space. We first determine optimal hyperparameters for each scheduler on a small number of epochs, fix these hyperparameters, and then evaluate performance as the number of epochs increases. Experimental results on various deep learning tasks (e.g., image classification, time series forecasting, and operator learning) demonstrate that both HyperbolicLR and ExpHyperbolicLR achieve more consistent performance improvements than conventional schedulers as training duration grows. These findings suggest that our hyperbolic-based schedulers offer a more robust and efficient approach to deep network optimization, particularly in scenarios constrained by computational resources or time.
\end{abstract}

\section{Introduction}
\label{sec:intro}

Recent steady progression of deep learning has prompted researchers to actively apply these techniques across diverse fields.
While there are several factors contributing to the success of deep learning, the advancement of optimization techniques is crucial, with learning rate scheduling now regarded as an almost essential process \cite{bengio2012practical, sun2019optimization, goyal2017accurate, you2017large}.
Learning rate scheduling is the process of appropriately changing the learning rate during training to enable more efficient learning.
Effective learning rate scheduling can significantly improve model convergence, generalization, and overall performance by adapting the learning process to different stages of training  \cite{smith2017cyclical,loshchilov2016sgdr}.

However, it also adds complexity by requiring the exploration of not only the model's hyperparameters but also the scheduler's hyperparameters.
This additional hyperparameter optimization becomes increasingly time-consuming and costly, especially for large neural network models.
Moreover, many existing learning rate schedulers exhibit sensitivity to the number of epochs, leading to what we term the \textit{learning curve decoupling problem}.
This problem manifests when the learning rate change pattern significantly differs upon altering only the total number of training epochs while keeping other hyperparameters fixed, resulting in inconsistent optimization behavior across different training durations.

This discrepancy can lead to suboptimal model performance and increased complexity in the hyperparameter optimization process, especially when scaling up training to larger epochs.
To address this epoch sensitivity issue, we focused on hyperbolic curves.
Unlike polynomial or trigonometric functions, hyperbolic curves have the unique property of converging to an asymptote as they move away from the vertex \cite{protter1970college}.
We expected that by utilizing this property, we could obtain a learning rate scheduler with higher flexibility, allowing small changes while not significantly altering the learning rate change pattern even as the number of epochs increases.
This approach aims to provide a more robust and adaptable learning rate scheduling method that maintains consistent performance across varying training durations.

The main contributions of this paper are as follows:
\begin{itemize}
   \item We propose HyperbolicLR and ExpHyperbolicLR, two novel learning rate schedulers based on hyperbolic curves that maintain a consistent initial learning rate change, regardless of the number of epochs.
   \item We provide detailed mathematical formulations and analyses of the proposed schedulers, highlighting their properties and advantages over existing methods.
   \item We conduct extensive experiments on various deep learning architectures and datasets to evaluate the performance of HyperbolicLR and ExpHyperbolicLR, comparing them to widely used learning rate schedulers.
   \item We make our code publicly available to facilitate further research and reproducibility.
\end{itemize}

\section{Related Work and Background}
\label{sec:related}

\subsection{Learning Rate Scheduling In Deep Learning}
\label{subsec:lr_scheduling}

The learning rate is a crucial hyperparameter in training deep learning models \cite{bengio2012practical,breuel2015effects,goodfellow2016deep}.
Learning rate scheduling dynamically adjusts this parameter during training to improve model performance and convergence.
Generally, it starts with a high learning rate and gradually decreases it, though some methods like cyclic learning rate \cite{smith2017cyclical}, warm restart \cite{loshchilov2016sgdr} or one cycle learning rate \cite{smith2019super} employ more complex patterns.

The significance of learning rate scheduling is multifaceted \cite{wu2023selecting}.
It enhances model accuracy while potentially reducing training time, improves adaptability during different phases of training, and increases training stability by preventing oscillation or divergence.
Furthermore, it allows the training process to adjust the learning speed and direction according to the needs of different phases, maintaining optimal learning efficiency and ensuring effective convergence.
By adjusting the learning rate according to the optimization landscape, learning rate scheduling enhances training stability, mitigating issues like oscillation or stagnation.

These benefits underscore why learning rate scheduling is now regarded as an almost essential process in deep learning optimization.
Given its impact on model performance, efficiency, and stability, developing effective learning rate scheduling techniques remains a key challenge in deep learning research.

\subsection{Common Learning Rate Schedulers}

Let $\mathbb{N}_0$ be the set of non-negative integers and $\mathcal{P}$ be the set of specific hyperparameters required for learning rate scheduling.
A learning rate scheduler can be expressed as a function $f$ mapping from $\mathbb{N}_0 \times \mathcal{P}$ to $[0, \infty)$ as follows:
\begin{equation}
\eta_n = f(n; P)
\end{equation}
Here, $n \in \mathbb{N}_0$ represents the current epoch, $P \in \mathcal{P}$ is a tuple of hyperparameters and $\eta_n$ is the learning rate at the current epoch.
Using this notation, we will describe some of the most widely used learning rate schedulers.

\begin{itemize}
   \item \textit{Polynomial learning rate scheduler} (PolynomialLR) \cite{chen2017deeplab,liu2015parsenet}:
         PolynomialLR decays the learning rate from the initial value using a polynomial function.
         \begin{equation}
            f(n; \eta_\text{init}, N, p) = \eta_\text{init} \times \left(1 - \frac{n}{N}\right)^p
         \end{equation}

   \item \textit{Cosine annealing learning rate scheduler} (CosineAnnealingLR) \cite{loshchilov2016sgdr}:
         CosineAnnealingLR decays the learning rate from an initial value $\eta_\text{max}$ to a minimum learning rate $\eta_\text{min}$ using a cosine curve.
         \begin{equation}
         \begin{split}
         f(n; \eta_\text{min}, \eta_\text{max}, N) = \eta_\text{min} + \frac{1}{2} (\eta_\text{max} - \eta_\text{min}) \\
                                                    \times (1 + \cos \left( \frac{n}{N} \pi\right ))
         \end{split}
         \end{equation}

   \item \textit{Exponential learning rate scheduler} (ExponentialLR) \cite{ioffe2015batch}:
         ExponentialLR decays the learning rate exponentially with a constant decay rate $\gamma$.
         Unlike the two schedulers explained earlier, this scheduler does not depend on the total number of epochs.
         \begin{equation}
         f(n; \eta_\text{init}, \gamma) = \eta_\text{init} \times \gamma^{n}
         \end{equation}
\end{itemize}

These learning rate schedulers are widely used in training deep learning models as they effectively improve model performance \cite{pytorch2024optim, papers2024code}.
However, each of these learning rate schedulers has its own set of problems.
In this study, we will particularly focus on the learning curve decoupling problem, a common issue shared by schedulers that depend on the total number of epochs, such as PolynomialLR and CosineAnnealingLR.

\subsection{Learning Curve Decoupling Problem}

This section introduces the learning curve decoupling problem, a critical issue in learning rate scheduling that has not been explicitly addressed in previous literatures. This problem arises when the learning rate change pattern significantly differs upon altering only the total number of training epochs while keeping other hyperparameters fixed. This leads to a separation of learning rate curves and, consequently, inconsistent optimization behavior, increasing the complexity of hyperparameter optimization.

To quantify this phenomenon, we propose the smoothed learning curve difference metric.
Consider a learning rate scheduler $f(n;N,P)$, where we fix other hyperparameters $P$ and change the total number of epochs from $N_1$ to $N_2$.
Let $l_1$ and $l_2$ be the loss or accuracy curves obtained from training with $N_1$ and $N_2$ epochs, respectively. The smoothed learning curve difference is defined as:
\begin{equation}
   \Delta_{\mathcal{S}}(l_1, l_2) = \frac{1}{N} \sum_{n=0}^{N-1} \frac{|\mathcal{S}(l_1)(n) - \mathcal{S}(l_2)(n)|}{\mathcal{S}(l_1)(n) + \mathcal{S}(l_2)(n)}
   \label{eq:smoothed_learning_curve_diff}
\end{equation}
Here, $\mathcal{S}$ is a smoothing operator, such as an exponential moving average \cite{brown1956exponential} or Savitzky-Golay filter \cite{savitzky1964smoothing}, and $N = \min(N_1, N_2)$.
The smoothing operator is introduced to mitigate the effect of noise that might falsely indicate decoupling when the underlying trends are actually similar. 
This metric approaches 0 for nearly identical curves and 1 for completely decoupled curves.

\begin{figure}[ht]
\vskip 0.1in
\begin{center}
\subfloat[CosineAnnealingLR]{
    \includegraphics[width=.47\columnwidth]{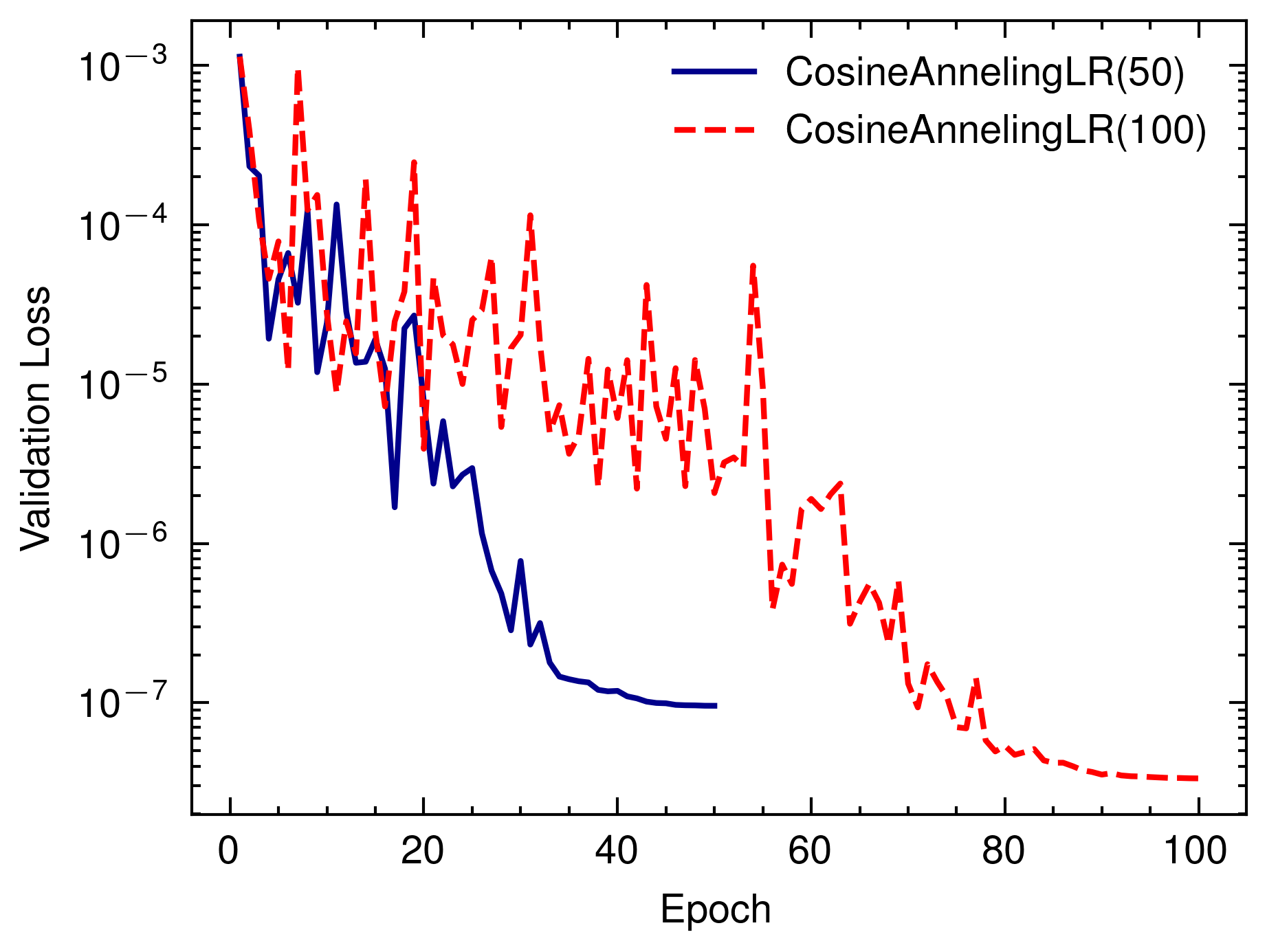}
}
\subfloat[ExpHyperbolicLR]{
    \includegraphics[width=.47\columnwidth]{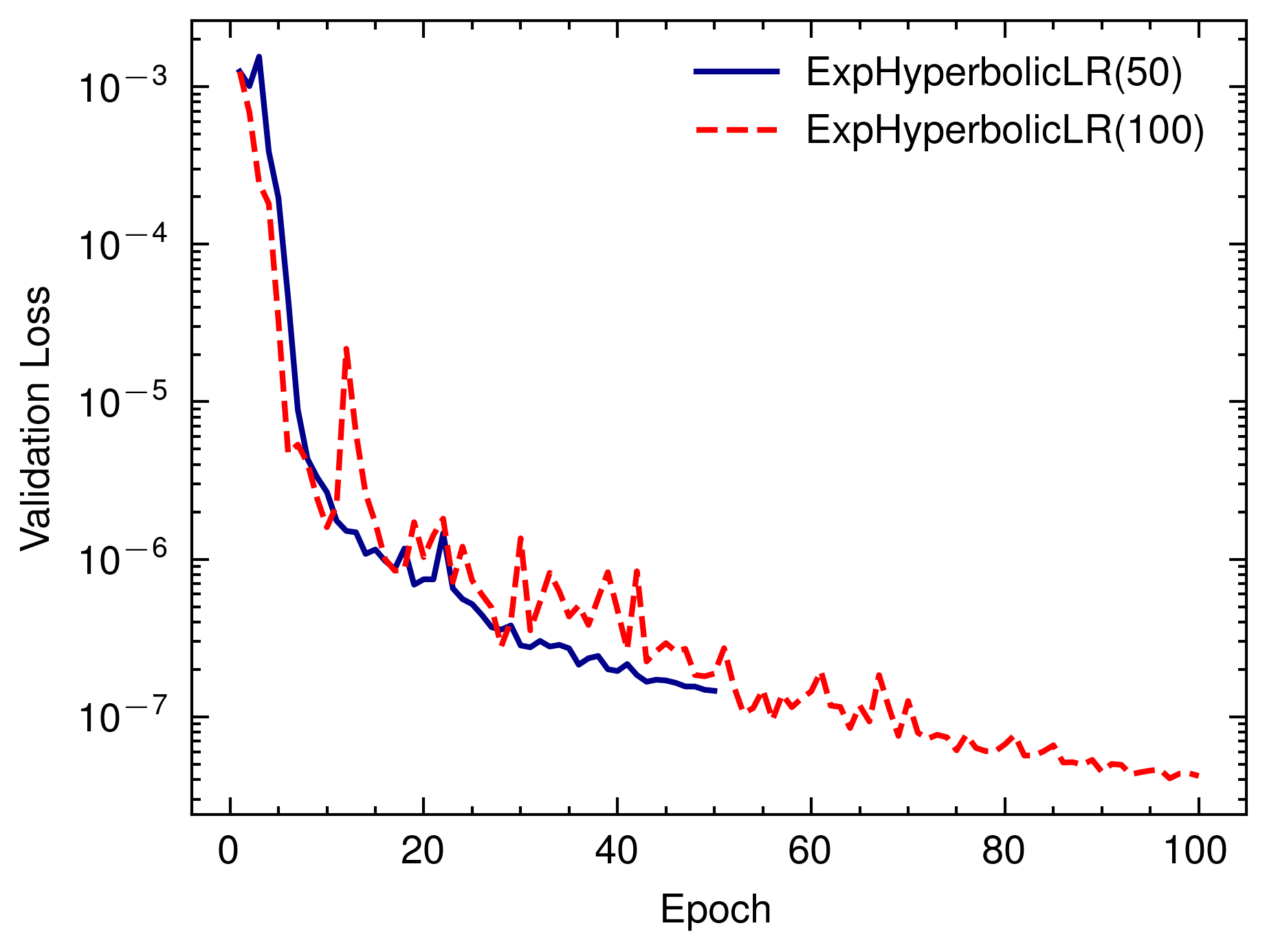}
}
\caption{
  Comparison of learning curves for CosineAnnealingLR (a) and ExpHyperbolicLR (b) in an time series forecasting task.
  Blue solid curves represent training with 50 epochs, while red dashed curves represent training with 100 epochs.
  Note the significant decoupling in CosineAnnealingLR after 20 epochs, compared to the more consistent behavior of ExpHyperbolicLR.
}
\label{fig:01_osc_learning_curve}
\end{center}
\vskip -0.1in
\end{figure}

To illustrate this problem, we conducted an experiment comparing CosineAnnealingLR and our proposed ExpHyperbolicLR in an time series forecasting task (Figure \ref{fig:01_osc_learning_curve}).
Both schedulers show similar trends up to about 20 epochs, after which CosineAnnealingLR exhibits significant decoupling of the learning curves, while ExpHyperbolicLR maintains a similar trend consistently.
Quantitatively, using the Savitzky-Golay filter, we obtain smoothed learning curve difference values of $6.7393 \times 10^{-1}$ for CosineAnnealingLR and $3.2157 \times 10^{-1}$ for ExpHyperbolicLR, confirming the visual observation.

The learning curve decoupling problem presents a significant challenge in optimizing deep learning models, particularly when adjusting training durations. Our study addresses this issue by proposing a novel approach based on hyperbolic curves, which we introduce in the following section.

\subsection{Hyperbolic Curves and Their Properties}
\label{subsec:hyperbolic_curves}

Hyperbolic curves are defined by the following equation:
\begin{equation}
\dfrac{x^2}{a^2} - \dfrac{y^2}{b^2} = 1
\end{equation}
where $a$ and $b$ are positive constants \cite{protter1970college}.
For learning rate scheduling, we focus on the branch where $x \leq -a$ and $y \geq 0$, represented by $y = \dfrac{b}{a} \sqrt{x^2 - a^2}$. 

A key property of hyperbolic curves, particularly relevant to our proposed schedulers, is the convergence of the curve's slope to its asymptotes as the distance from the vertex increases.
This can be demonstrated by examining the derivative:

\begin{equation}
   \frac{\mathrm{d}y}{\mathrm{d}x} = \frac{b}{a} \frac{x}{\sqrt{x^2 - a^2}}
\end{equation}

As $x$ approaches negative infinity, this derivative converges to $-b/a$:

\begin{equation}
   \lim_{x \rightarrow -\infty} \frac{\mathrm{d}y}{\mathrm{d}x} = -\frac{b}{a}
\end{equation}

This asymptotic behavior suggests that for $x \ll -a$, the rate of change of the curve becomes stable and predictable.
This property forms the basis for developing learning rate schedulers that maintain a consistent initial learning rate change, regardless of the total number of epochs.

\section{Proposed Learning Rate Schedulers}
\label{sec:proposed}

\subsection{HyperbolicLR}
\label{subsec:hyperboliclr}

We introduce HyperbolicLR, a novel learning rate scheduler based on the hyperbolic curve.
The learning rate at epoch $n$ is defined as:
\begin{multline}
   f_\text{H}(n; \eta_\text{init}, \eta_\text{inf}, N, U) = \\ \eta_\text{init} + (\eta_\text{init} - \eta_\text{inf}) (h(n; N, U) - h(0; N, U))
   \label{eq:hyperbolic_lr}
\end{multline}
where $\eta_\text{init}$ denotes the initial learning rate, $\eta_\text{inf}$ represents the infimum of the learning rate, $N$ is the total number of epochs minus one, and $U$ is the upper bound of $N$.
The function $h(n; N, U)$ is given by:
\begin{equation}
   h(n; N, U) = \sqrt{\frac{N - n}{U} \left(2 - \frac{N + n}{U}\right)} \quad (U \geq N)
\end{equation}
This function represents a portion of a hyperbolic curve with vertex at $(N, 0)$, center at $(U, 0)$, and asymptote slope of $-1/U$.
Thus, HyperbolicLR starts at $\eta_\text{init}$ and decreases along the hyperbolic curve, ending at a learning rate that is always greater than or equal to $\eta_\text{inf}$ at the $N$-th epoch. (For proof, see section \ref{sec:props_and_proofs}).
The variable $N$ can be adjusted within the range less than or equal to $U$. 
When $U$ and $N$ are equal, the curve coincides with the hyperbola's asymptote, resulting in a simple linear decay.

A key property of HyperbolicLR is that the asymptote slope $-1/U$ does not depend on $N$, which corresponds to the total number of epochs.
Therefore, in the early stages of training when $n \ll N$, the change rate of the learning rate will be similar to the asymptote, meaning it won't change significantly even if $N$ is altered.
This characteristic suggests that HyperbolicLR could potentially address the learning curve decoupling problem by maintaining a consistent learning pattern across varying epoch numbers.

\subsection{ExpHyperbolicLR}
\label{subsec:exphyperboliclr}

Building upon the concept of HyperbolicLR, we propose ExpHyperbolicLR, which extends the hyperbolic approach to the exponential space.
The learning rate at epoch $n$ for ExpHyperbolicLR is defined as:
\begin{multline}
   f_\text{EH}(n; \eta_\text{init}, \eta_\text{inf}, N, U) =  \\ \eta_\text{init} \times \exp{\left(\ln\frac{\eta_\text{init}}{\eta_\text{inf}} \times (h(n; N, U) - h(0; N, U))\right)}
\end{multline}
where the hyperparameters are consistent with those in HyperbolicLR.
Notably, ExpHyperbolicLR can be expressed in terms of HyperbolicLR:
\begin{equation}
   f_\text{EH}(n; \eta_\text{init}, \eta_\text{inf}, N, U) = \exp{\left(f_\text{H}(n; \ln \eta_\text{init}, \ln \eta_\text{inf}, N, U)\right)}
   \label{eq:exphyp_as_hyp}
\end{equation}
The monotonically increasing nature of the exponential function ensures that ExpHyperbolicLR retains the fundamental properties of HyperbolicLR.
However, when $N = U$, it becomes a linear function in the exponential space, equivalent to ExponentialLR rather than exhibiting linear decay.
ExpHyperbolicLR decreases the learning rate exponentially, making its initial rate of decrease faster than HyperbolicLR.
This could make it a better choice in environments where overfitting occurs easily.

\subsection{Comparison with Commonly Used Schedulers}
\label{subsec:comparison}

\begin{figure}[ht]
\vskip 0.1in
\begin{center}
\subfloat[PolynomialLR]{
    \includegraphics[width=0.47\columnwidth]{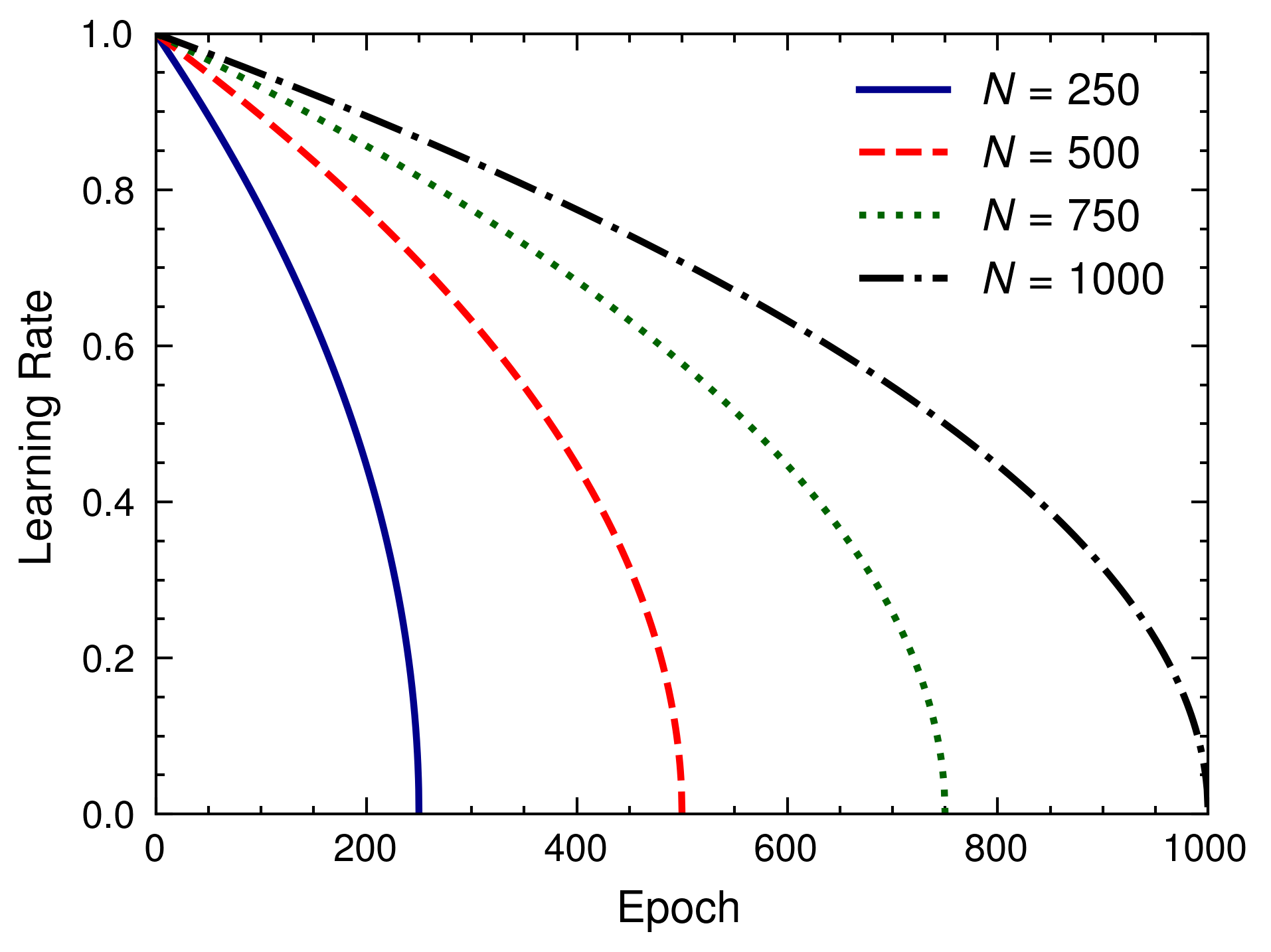}
}
\hfill
\subfloat[CosineAnnealingLR]{
    \includegraphics[width=0.47\columnwidth]{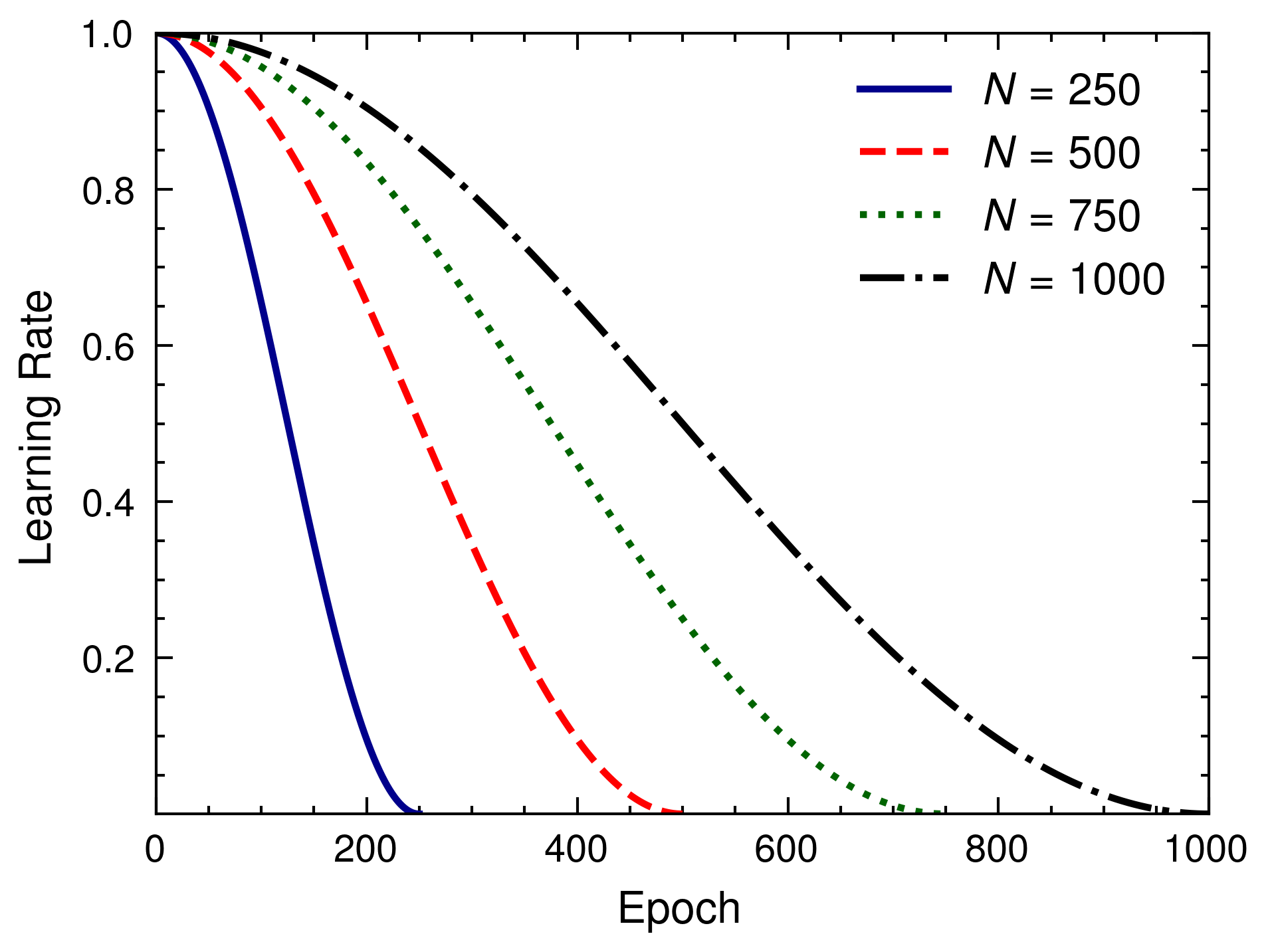}
}

\subfloat[HyperbolicLR]{
    \includegraphics[width=0.47\columnwidth]{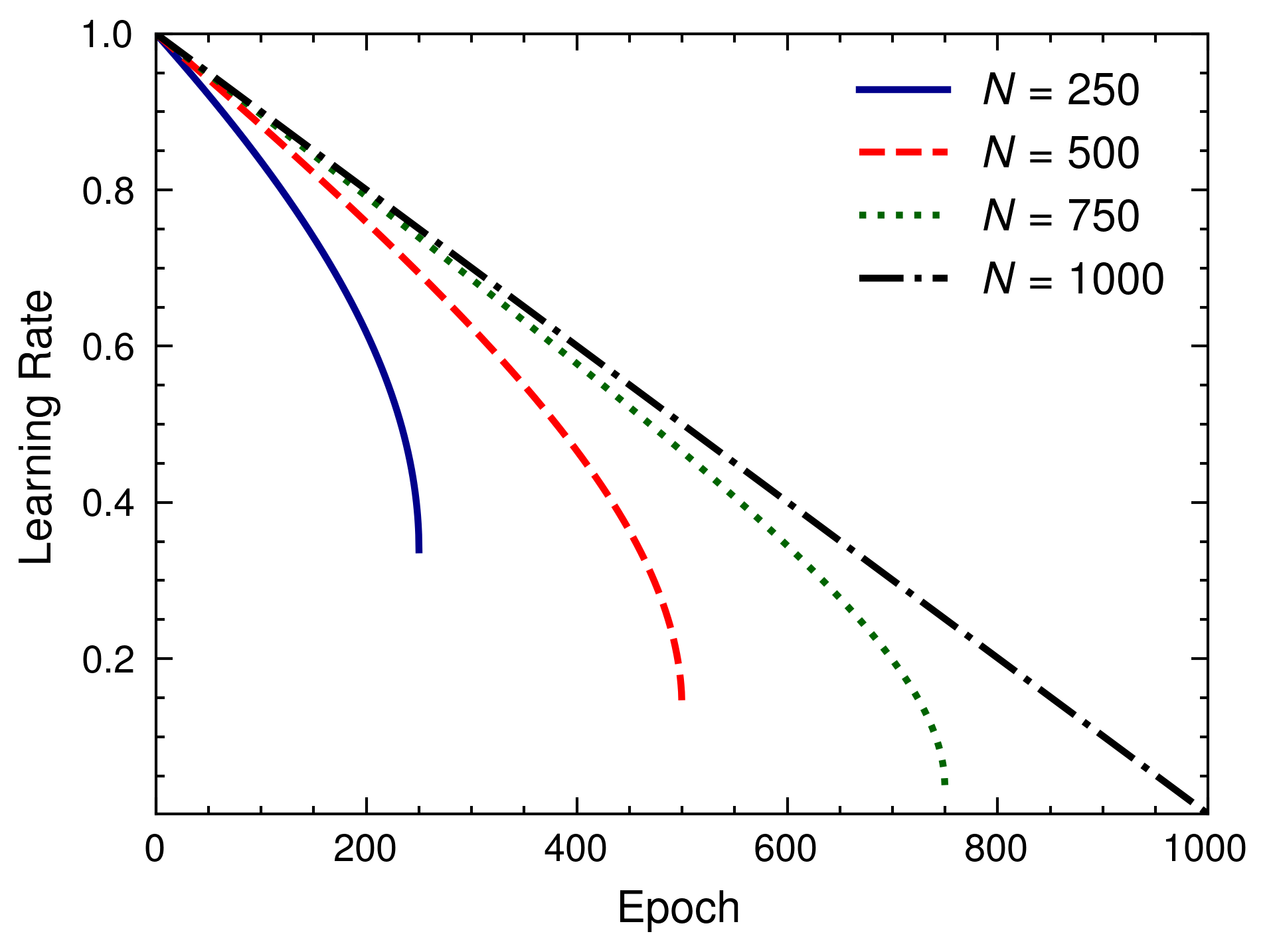}
}
\hfill
\subfloat[ExpHyperbolicLR]{
    \includegraphics[width=0.47\columnwidth]{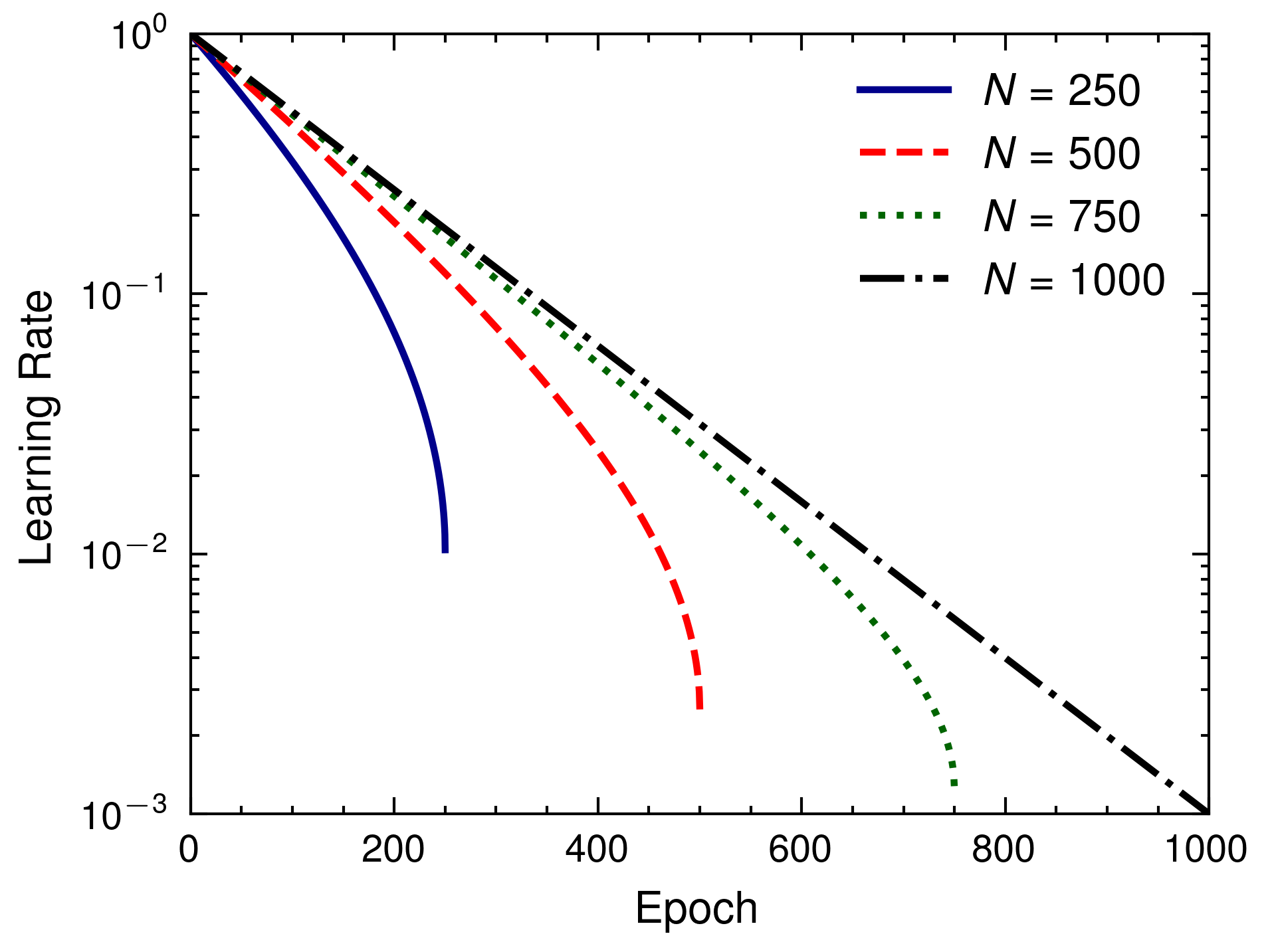}
}
\caption{
    Comparison of learning rate schedules for different total epochs ($N = 250,\,500,\,750,\,1000$). 
    The schedulers shown are:
    (a) PolynomialLR ($p = 0.5$), 
    (b) CosineAnnealingLR ($\eta_\text{min} = 10^{-4}$),
    (c) HyperbolicLR ($\eta_\text{inf} = 10^{-4},\,U = 1000$),
    and (d) ExpHyperbolicLR ($\eta_\text{inf} = 10^{-4},\,U = 1000$).
    All schedulers start with $\eta_\text{init} = 1$.
}
\label{fig:lr_schedulers_comparison}
\end{center}
\vskip -0.1in
\end{figure}

Figure \ref{fig:lr_schedulers_comparison} illustrates the behavior of PolynomialLR, CosineAnnealingLR, and our proposed HyperbolicLR and ExpHyperbolicLR schedulers for different total epoch settings ($N = 250, 500, 750, 1000$).
As evident from the figure, both PolynomialLR and CosineAnnealingLR exhibit significant variations in their learning rate decay patterns as the total number of epochs changes. In contrast, our proposed HyperbolicLR and ExpHyperbolicLR demonstrate more consistent decay patterns across different epoch settings, particularly in the early stages of training.

To quantify the initial learning rate change, we introduce the \textit{Initial Learning Rate Integral} (ILRI) metric:
\begin{equation}
   \text{ILRI} = \int_0^{n_{0.8}} \left|\eta_n - 0.8 \times \eta_\text{init} \right| \mathrm{d}n
\end{equation}
where $\eta_n$ is the learning rate at epoch $n$, and $n_{0.8}$ is the epoch at which the learning rate reaches $80\%$ of its initial value.
We use cubic Hermite splines to interpolate and treat epochs as real numbers for more precise measurements.
This focus on the initial learning rate changes aligns with observations that many important aspects of neural network learning occur within the earliest iterations or epochs of training \cite{frankle2020early,kalra2024phase}.

Table \ref{tab:ilri} presents the relative ILRI percentage difference for each scheduler with $N$ set to 250, 500, and 750, using $N=1000$ as a baseline. For HyperbolicLR and ExpHyperbolicLR, we set $\eta_\text{init}=1, \eta_\text{inf}=10^{-3},$ and $U=1000$. HyperbolicLR and ExpHyperbolicLR demonstrate remarkably lower percentage differences compared to PolynomialLR and CosineAnnealingLR, especially for higher epoch counts.

\begin{table}[t]
\caption{Percentage difference in Initial Learning Rate Integral (ILRI) for each learning rate scheduler compared to $N=1000$, with $N$ set to 250, 500, and 750.}
\label{tab:ilri}
\vskip 0.1in
\begin{center}
\begin{small}
\begin{sc}
\begin{tabular}{ccccc}
\toprule
LR Schedulers & $N=250$ & $N=500$ & $N=750$ \\
\midrule
PolynomialLR & $75\%$ & $50\%$ & $25\%$ \\
CosineAnnealingLR & $75\%$ & $50\%$ & $25\%$ \\
HyperbolicLR & $38.01\%$ & $15.31\%$ & $3.66\%$ \\
ExpHyperbolicLR & $34.46\%$ & $13.67\%$ & $3.24\%$ \\
\bottomrule
\end{tabular}
\end{sc}
\end{small}
\end{center}
\vskip -0.1in
\end{table}

These results indicate that HyperbolicLR and ExpHyperbolicLR maintain more consistent initial learning rate changes across varying epoch numbers, potentially leading to more stable training processes.

\section{Experimental Setup}
\label{sec:exp_setup}

To empirically validate our theoretical postulations and comprehensively evaluate the efficacy of the proposed learning rate schedulers, we designed a series of experiments across diverse deep learning tasks. This section outlines our experimental framework, including the datasets used, the selected model architectures, and our rigorous evaluation protocol.

\subsection{Datasets}
\label{subsec:datasets}

\subsubsection{CIFAR-10 for Image Classification}
\label{subsubsec:cifar10}

We employed the CIFAR-10 dataset \cite{krizhevsky2009learning} for our image classification experiments.
This dataset comprises 60,000 $32 \times 32$ color images across 10 classes, with 50,000 for training and 10,000 for validation.
We applied common data augmentation techniques to the training set, including random horizontal flips with a probability of 0.5, random crops to $32\times 32$ after padding by 4 pixels on each side, and normalization using the dataset's mean and standard deviation.

\subsubsection{Oscillation Dataset for Time Series Prediction}
\label{subsubsec:oscillation}

We created a custom dataset incorporating simple and damped harmonic oscillations, generated using an ordinary differential equation with varying damping ratios.
The dataset consists of 29,646 input-output pairs for time series prediction, with each input sequence comprising 100 consecutive time steps used to predict the subsequent 20 time steps.

\subsubsection{Integral Dataset for Operator Learning}
\label{subsubsec:integral}

For the operator learning task, we generated a dataset to learn an integral operator using a Gaussian Random Field approach with varying length scales.
The dataset consists of 10,000 functions sampled at 100 points each, along with 100 target points for operator evaluation.

Detailed descriptions of the dataset generation methods are provided in section \ref{sec:dataset_generation}.

\subsection{Model Architectures}
\label{subsec:models}

We employed four different model architectures for our experiments:

\begin{itemize}
   \item SimpleCNN for CIFAR-10 classification
   \item LSTM Sequence-to-Sequence (LSTM Seq2Seq) model \cite{sutskever2014sequence} for oscillation prediction
   \item DeepONet \cite{lu2021learning} for learning integral operator
   \item TraONet, a novel transformer-based operator network, also for learning integral operator
\end{itemize}

Each architecture was chosen to suit its respective task.
SimpleCNN uses convolutional layers \cite{krizhevsky2012imagenet} for feature extraction, followed by fully connected layers for classification.
The LSTM Seq2Seq model employs an encoder-decoder structure for time series prediction.
DeepONet and TraONet are designed for operator learning, with TraONet incorporating transformer \cite{vaswani2017attention} components for enhanced performance.
Detailed descriptions and illustrations of these architectures are provided in section \ref{sec:model_architectures}.

\subsection{Experimental Design and Evaluation Protocol}
\label{subsec:exp_design}

Our experimental protocol, designed to evaluate various learning rate schedulers across diverse tasks and model architectures, comprised three phases:

\begin{enumerate}
   \item \textbf{Model Hyperparameter Optimization}: We optimized model hyperparameters using grid search with a fixed scheduler.
   \item \textbf{Scheduler Hyperparameter Optimization}: Using the optimized model hyperparameters, we then optimized scheduler hyperparameters over 50 epochs using the Tree-structured Parzen Estimator \cite{bergstra2011algorithms}.
   \item \textbf{Performance Evaluation}: We assessed each scheduler's performance using the optimized hyperparameters, progressively increasing the number of epochs from 50 to 200 in increments of 50.
\end{enumerate}

We employed the AdamW optimizer \cite{loshchilov2017decoupled} with $\beta_1 = 0.9, \beta_2 = 0.999, \epsilon=10^{-8}$ and weight decay $\lambda = 0.01$ across all experiments.
Batch sizes were set to 256 for CIFAR-10 and oscillation prediction tasks, and 100 for the operator learning task.
Each training session was conducted five times using different random seeds (89, 231, 928, 814, 269) obtained from random.org \cite{RandomOrg}.

To analyze performance, we calculated relative performance improvements per 50-epoch interval.
We also utilized the previously introduced smoothed learning curve difference metric (Eq. \ref{eq:smoothed_learning_curve_diff}) to quantify learning curve decoupling.

Detailed information on hyperparameter optimization ranges, optimal hyperparameters, technical setup, and learning curves for each scheduler across all experiments are provided in the section \ref{sec:experimental_design} and \ref{sec:learning_curves}.

\section{Results and Analysis}
\label{sec:results}

\subsection{Overall Performance}
\label{subsec:overall_performance}

\begin{figure}[ht]
\vskip 0.1in
\begin{center}
    \subfloat[CNN (left) and LSTM (right)]{
   \includegraphics[width=\columnwidth]{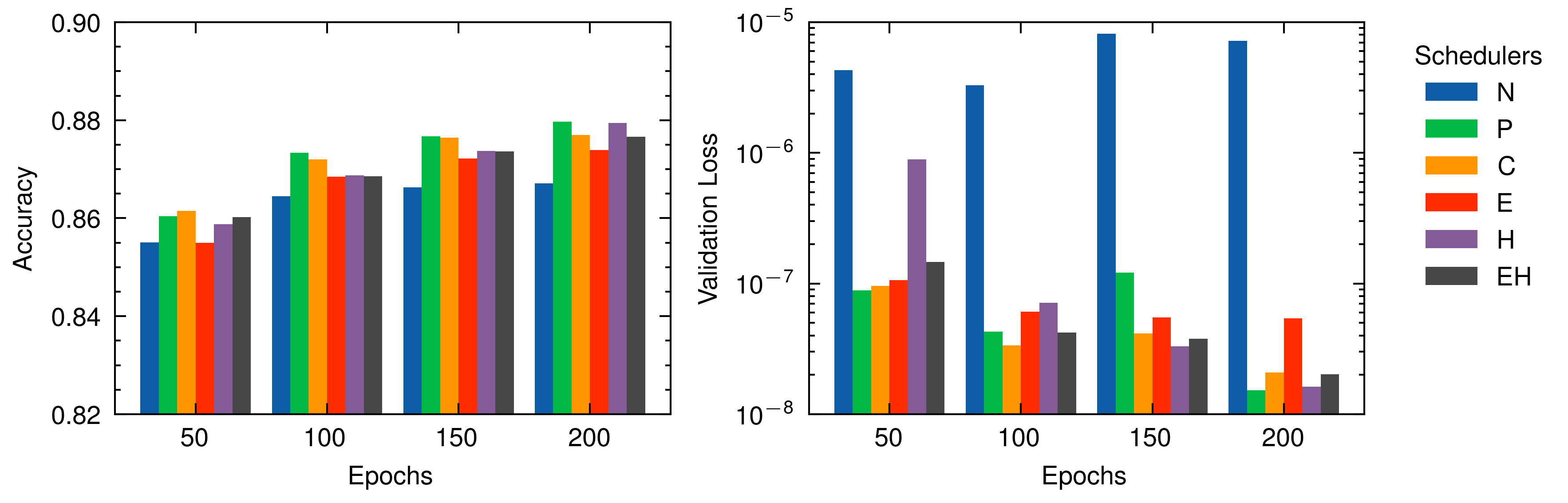}
}

    \subfloat[DeepONet (left) and TraONet (right)]{
   \includegraphics[width=\columnwidth]{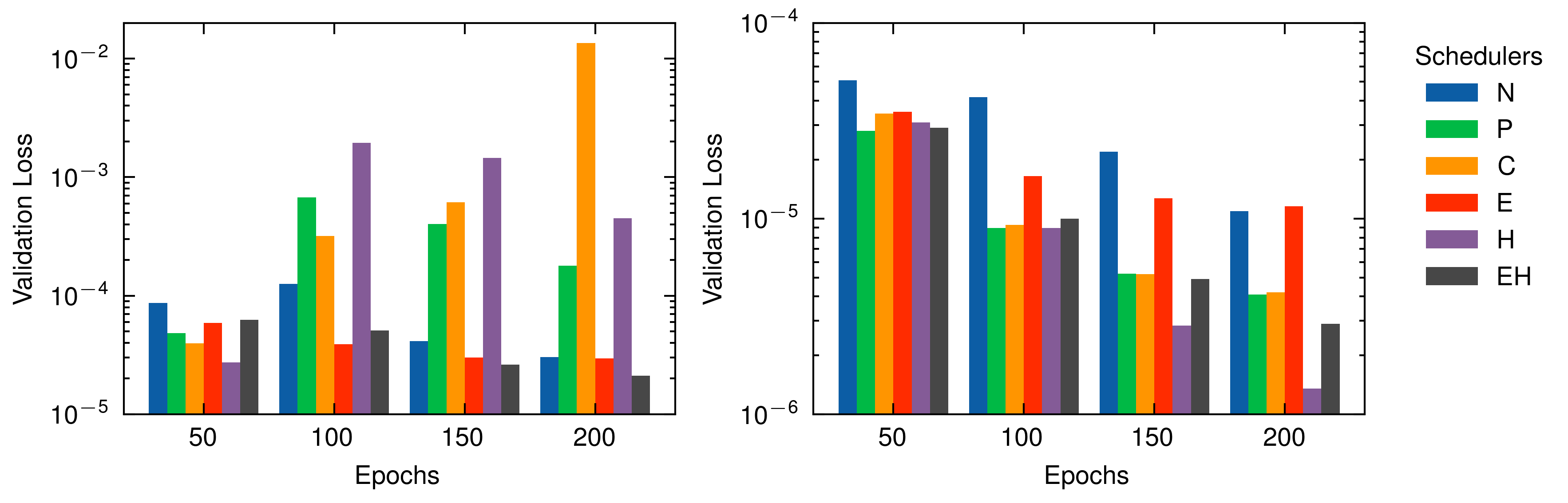}
}
\caption{
   Performance of all schedulers for all tasks and architectures.
   Scheduler abbreviations: N = No scheduler (constant learning rate), P = PolynomialLR, C = CosineAnnealingLR, E = ExponentialLR, H = HyperbolicLR, EH = ExpHyperbolicLR.
}
\label{fig:overall_performance}
\end{center}
\vskip -0.1in
\end{figure}

Figure \ref{fig:overall_performance} presents the performance of all schedulers across the four model architectures studied.
In nearly all cases, the use of learning rate schedulers led to improved performance compared to using no scheduler, underscoring the importance of learning rate scheduling in optimizing deep learning models.

For the SimpleCNN model on CIFAR-10, PolynomialLR consistently outperformed other schedulers. However, HyperbolicLR exhibited a sharp performance increase at 200 epochs, nearly matching PolynomialLR's performance at this point.

In the LSTM Seq2Seq model for time series prediction, HyperbolicLR and ExpHyperbolicLR displayed steady improvements across all epochs, with HyperbolicLR showing the most pronounced performance gains.

The operator learning task using DeepONet revealed an interesting phenomenon where non-exponential decay models experienced significant loss divergence after 100 epochs.
Conversely, ExponentialLR and ExpHyperbolicLR demonstrated consistent performance improvements, with ExpHyperbolicLR achieving the best performance from 150 epochs onward.

For the TraONet model, HyperbolicLR emerged as the top performer after 100 epochs, with ExpHyperbolicLR consistently ranking second in performance.

These results suggest that transformer-based architectures may be more resilient to scheduler choice, though still benefiting from optimized scheduling strategies.

\subsection{Performance Metrics and Learning Curve Analysis}
\label{subsec:performance_metrics}

\begin{table*}[!t]
\caption{
   Performance metrics and power regression results across all models and schedulers.
   Scheduler abbreviations are the same as in Figure \ref{fig:overall_performance}.
   $\mu$ and $\sigma$ represent the mean and standard deviation of performance improvement.
   $B$, $R^2$, and $p$-value are from power regression analysis ($y = \exp(A) x^B$).
   sLCD (Smoothed Learning Curve Difference) indicates the degree of learning curve decoupling, with lower values being better.
   For DeepONet, only $\mu$ and $\sigma$ were calculated due to learning instability. Dashes (---) indicate zero values. Best values among schedulers are in \textbf{bold}, second-best are \underline{underlined}.}
\label{tab:combined-performance-metrics}
\vskip 0.1in
\begin{center}
\begin{small}
\begin{sc}
\begin{tabular}{ccccccccc}
\toprule
\textbf{Model} & \textbf{Index} & \textbf{N} & \textbf{P} & \textbf{C} & \textbf{E} & \textbf{H} & \textbf{EH} \\
\midrule
\multirow{6}{*}{\textbf{SimpleCNN}} & $\mu$ (\%) & 0.46 & \underline{0.74} & 0.60 & 0.73 & \textbf{0.79} & 0.63 \\
& $\sigma$ (\%) & 0.55 & 0.67 & 0.58 & 0.74 & \textbf{0.31} & \underline{0.32} \\
& $B$ & 0.0102 & 0.0160 & 0.0134 & \underline{0.0161} & \textbf{0.0167} & 0.0137 \\
& $R^2$ & 0.9139 & 0.9639 & 0.9543 & 0.9467 & \underline{0.9949} & \textbf{0.9990} \\
& $p$-value & 0.0440 & 0.0182 & 0.0231 & 0.0270 & \underline{0.0026} & \textbf{0.0008} \\
& sLCD ($\times 10^{-3}$) & --- & 1.04 & 2.68 & --- & \underline{0.902} & \textbf{0.748} \\
\midrule
\multirow{6}{*}{\textbf{LSTM Seq2Seq}} & $\mu$ (\%) & -37.57 & -14.20 & 30.22 & 17.84 & \textbf{65.51} & \underline{42.65} \\
& $\sigma$ (\%) & 95.42 & 146.1 & 47.61 & \textbf{21.80} & \underline{22.96} & 30.79 \\
& $B$ & 0.4879 & -0.8070 & -0.9662 & -0.4982 & \textbf{-2.8715} & \underline{-1.3431} \\
& $R^2$ & 0.4688 & 0.2782 & 0.8342 & 0.8882 & \textbf{0.9799} & \underline{0.9493} \\
& $p$-value & 0.3153 & 0.4726 & 0.0867 & 0.0576 & \textbf{0.0101} & \underline{0.0257} \\
& sLCD & --- & 0.553 & 0.613 & --- & \underline{0.331} & \textbf{0.254} \\
\midrule
\multirow{2}{*}{\textbf{DeepONet}} & $\mu$ (\%) & 16.26 & -401.7 & -970.5 & \underline{19.56} & -2304 & \textbf{28.83} \\
& $\sigma$ (\%) & 56.74 & 779.0 & 1034 & \textbf{16.23} & 4073 & \underline{17.06} \\
\midrule
\multirow{6}{*}{\textbf{TraONet}} & $\mu$ (\%) & 38.59 & 43.85 & 45.44 & 28.37 & \textbf{63.87} & \underline{52.50} \\
& $\sigma$ (\%) & 17.76 & 23.20 & 26.89 & 22.60 & \textbf{10.28} & \underline{12.45} \\
& $B$ & -1.0576 & -1.4159 & -1.5561 & -0.8229 & \textbf{-2.2630} & \underline{-1.6585} \\
& $R^2$ & 0.8344 & 0.9877 & 0.9789 & 0.9601 & \underline{0.9878} & \textbf{0.9983} \\
& $p$-value & 0.0865 & 0.0062 & 0.0106 & 0.0202 & \underline{0.0061} & \textbf{0.0008} \\
& sLCD & --- & 0.151 & 0.231 & --- & \underline{0.117} & \textbf{0.0446} \\
\bottomrule
\end{tabular}
\end{sc}
\end{small}
\end{center}
\vskip -0.1in
\end{table*}

Table \ref{tab:combined-performance-metrics} presents a comprehensive overview of performance metrics across all models and schedulers.
This includes the mean ($\mu$) and standard deviation ($\sigma$) of performance improvement, the smoothed Learning Curve Difference (sLCD), and power regression analysis results ($B$ coefficient, $R^2$ value, and $p$-value).
These metrics collectively provide insights into the consistency of improvement, learning curve stability, and performance trends over extended training periods for each scheduler.

\subsubsection{Consistency of Improvement}
\label{subsubsec:consistency}

The consistency of improvement is reflected in the mean ($\mu$) and standard deviation ($\sigma$) of performance gains across epochs. For SimpleCNN, HyperbolicLR demonstrated the highest mean improvement (0.79\%) while maintaining the lowest standard deviation (0.31\%), indicating both superior performance enhancement and consistency. ExpHyperbolicLR showed comparable consistency (0.32\% std dev) but with a lower mean improvement (0.63\%).

In the LSTM Seq2Seq task, HyperbolicLR significantly outperformed other schedulers with a mean improvement of 65.51\%. ExpHyperbolicLR ranked second with 42.65\%. Notably, ExponentialLR showed the lowest standard deviation (21.80\%), suggesting very consistent, albeit slower, improvement.

The DeepONet task revealed unique challenges, with ExpHyperbolicLR being the only scheduler to show positive mean improvement (28.83\%) and low variability (17.06\% std dev). Other schedulers, except ExponentialLR, exhibited negative mean improvements, indicating potential instability in the learning process.

For TraONet, HyperbolicLR again led in mean improvement (63.87\%) and consistency (10.28\% std dev), followed closely by ExpHyperbolicLR (52.50\% mean, 12.45\% std dev).

These results demonstrate that HyperbolicLR and ExpHyperbolicLR consistently provide the most substantial and stable performance improvements across diverse tasks and model architectures.

\subsubsection{Power Regression Analysis}
\label{subsubsec:power_regression}

To better understand performance trends, we conducted power regression analysis on the relationship between epoch numbers and validation metrics, using the model $y = \exp(A) x^B$. Table \ref{tab:combined-performance-metrics} presents the $B$ coefficient, $R^2$ value, and $p$-value for each scheduler-model combination.

HyperbolicLR and ExpHyperbolicLR consistently showed the highest $R^2$ values and lowest $p$-values across models, indicating the most consistent and predictable performance improvements. HyperbolicLR exhibited the highest absolute $B$ values, suggesting the steepest improvement rate over epochs.

ExponentialLR demonstrated statistically significant regressions but with lower $B$ values, indicating slower improvement over time. PolynomialLR and CosineAnnealingLR showed variable performance, with strong results for SimpleCNN and TraONet but poor performance for LSTM Seq2Seq, suggesting task sensitivity.

The no-scheduler baseline only showed significant regression for SimpleCNN, indicating unpredictable performance improvements for other models without scheduling.

For the CIFAR-10 task with SimpleCNN, we used accuracy for regression due to label smoothing effects on the loss function. The DeepONet model's regression analysis was omitted due to learning instability, as reflected in its highly variable $\mu$ and $\sigma$ values.

\subsubsection{Learning Curve Decoupling Analysis}
\label{subsubsec:learning_curve_decoupling}

The smoothed Learning Curve Difference (sLCD) quantifies the extent of learning curve decoupling, with lower values indicating more consistent learning curves across different epoch settings.

ExpHyperbolicLR consistently demonstrated the lowest sLCD values across all tasks: $7.48 \times 10^{-4}$ for SimpleCNN, 0.254 for LSTM Seq2Seq, and 0.0446 for TraONet. HyperbolicLR followed closely with $9.02 \times 10^{-4}$, 0.331, and 0.117 for the respective models. 

PolynomialLR and CosineAnnealingLR exhibited higher sLCD values, indicating greater decoupling. For SimpleCNN, PolynomialLR showed an sLCD of $1.04 \times 10^{-3}$, while CosineAnnealingLR had $2.68 \times 10^{-3}$. This trend persisted across other models, with PolynomialLR and CosineAnnealingLR showing values of 0.553 and 0.613 for LSTM Seq2Seq, and 0.151 and 0.231 for TraONet, respectively.

No scheduler and ExponentialLR show zero sLCD by definition due to their non-adaptive nature. However, this comes at the cost of potentially suboptimal learning rate adjustments during training.

\subsubsection{Summary}
\label{subsubsec:summary}

In summary, our comprehensive evaluation revealed that HyperbolicLR and ExpHyperbolicLR consistently demonstrated superior performance across various deep learning tasks and model architectures.
These schedulers showed more consistent performance improvements and better stability across increasing epoch numbers compared to traditional schedulers.
The power regression analysis further supported these findings, with HyperbolicLR and ExpHyperbolicLR exhibiting the highest $R^2$ values and lowest $p$-values, indicating the most consistent and predictable performance improvements over extended training periods.
Additionally, HyperbolicLR often showed the highest absolute $B$ values, suggesting the steepest improvement rate over epochs.

Furthermore, our proposed schedulers exhibited a superior ability to maintain stable learning curves across different epoch settings, as evidenced by lower smoothed Learning Curve Difference (sLCD) values.
This stability potentially leads to more reliable and efficient training when scaling to longer durations. 

While traditional schedulers like PolynomialLR and CosineAnnealingLR showed good performance in some scenarios, they often exhibited less consistent improvements, higher learning curve decoupling, and more variable regression results across different tasks.
These findings suggest that our proposed hyperbolic-based learning rate schedulers offer a promising approach to improving the training of deep neural networks, particularly in scenarios where consistent and predictable performance across varying epoch settings is crucial.

\section{Conclusion}
\label{sec:conclusion}

This study introduced HyperbolicLR and ExpHyperbolicLR, novel learning rate schedulers designed to address the learning curve decoupling problem. These schedulers demonstrated superior stability and predictability across varying epoch settings, as evidenced by lower smoothed learning curve difference values and favorable power regression analysis results.

The key contribution of our work lies in the development of schedulers that maintain consistent performance improvements as training duration increases. This property is particularly valuable in scenarios where computational resources limit extensive hyperparameter searches for different training durations. By providing more reliable scaling to longer training periods, our approach could significantly reduce the time and resources required for hyperparameter optimization in deep learning tasks.

However, it's important to acknowledge the limitations of our study. While our schedulers showed promising results across various tasks and architectures, the optimal choice of scheduler may still depend on specific learning scenarios. Future research should explore the performance of these schedulers in a wider range of contexts, including more diverse optimization algorithms, network architectures, and application domains such as natural language processing or reinforcement learning.

Additionally, further investigation into the theoretical foundations of hyperbolic-based learning rate scheduling could provide deeper insights into their effectiveness and potentially lead to even more robust scheduling techniques. Exploring the interaction between these schedulers and other optimization techniques, such as adaptive gradient methods or curriculum learning, could also yield interesting findings.

In conclusion, HyperbolicLR and ExpHyperbolicLR represent a significant step towards more efficient and reliable deep learning optimization. While they offer a promising approach to improving training consistency and predictability, they should be viewed as valuable additions to the deep learning toolkit rather than universal replacements for existing methods. As deep learning continues to evolve, the development of such specialized tools will play a crucial role in advancing the field and expanding its applications.

\section*{Impact Statement}

Our proposed learning rate schedulers, HyperbolicLR and ExpHyperbolicLR, primarily aim to advance the technical capabilities of deep learning optimization. While the direct ethical implications may be limited, we identify several potential broader impacts:

\begin{enumerate}
    \item \textit{Environmental Impact:} By providing more consistent and predictable training behavior across different epoch settings, our schedulers may reduce the need for multiple training runs and hyperparameter searches, potentially leading to reduced computational resources and associated energy consumption.

    \item \textit{Accessibility:} The improved stability and predictability of our schedulers could make deep learning more accessible to researchers and practitioners with limited computational resources, particularly in developing regions or smaller organizations.

    \item \textit{Research Efficiency:} More reliable learning rate scheduling could accelerate research progress across various domains where deep learning is applied, including healthcare, climate science, and scientific discovery.
\end{enumerate}

However, like any advancement in machine learning optimization, these techniques could also enable both beneficial and potentially harmful applications. We encourage users to consider the ethical implications of their specific use cases and implement appropriate safeguards.

\nocite{langley00}

\bibliography{ref}
\bibliographystyle{icml2025}

\newpage
\appendix
\onecolumn

\section{Propositions and Proofs}
\label{sec:props_and_proofs}

\begin{proposition}
Let $h$ be a function defined by
\begin{equation}
    h(n; N, U) = \sqrt{\frac{N - n}{U} \left(2 - \frac{N + n}{U}\right)} \quad (U \geq N)
    \label{eq:hyperbolic}
\end{equation}
Then the graph $\{(n, h(n; N, U)) | 0 \leq n \leq N < U\}$ represents the part of hyperbolic curve.

\end{proposition}

\noindent \textit{Proof}
Let us denote $x = n$, $y = h(n;N,U)$, and expand Equation \ref{eq:hyperbolic}. We can then obtain the following equation:
\begin{equation}
\frac{(x - U)^2}{(U - N)^2} - \frac{y^2}{((U - N) / U)^2} = 1 \quad (0 \leq x \leq N, y \geq 0).
\label{eq:hyperbola}
\end{equation}
This equation represents a standard form of a hyperbola centered at $(U, 0)$ with the transverse axis along the $x$-axis.
The semi-major axis has a length of $U-N$, and the asymptotes have slopes of $-1/U$.
It is important to note that when $N=U$, the function $h(n;N,N)$ becomes an asymptote. \hfill$\Box$

\begin{proposition}
For all $N, U \in \mathbb{R}$ with $0 < N \leq U$, we have $h(0; N, U) \leq 1$. Equality holds if and only if $N = U$.
\end{proposition}

\noindent \textit{Proof}
Recall that $h(0; N, U) = \sqrt{N/U (2 - N/U)}$ for $0 < N \leq U$.
Let $x = N/U$. Then $0 < x \leq 1$, and we can rewrite $h(0; N, U)$ as:
\[
h(0; N, U) = \sqrt{x(2-x)}
\]
Since $x$ and $2-x$ are non-negative numbers, by the inequality of arithmetic and geometric means (AM-GM inequality), we have:
\[
h(0; N, U) = \sqrt{x(2-x)} \leq \frac{x + (2-x)}{2} = 1
\]
Equality holds if and only if $x = 2-x$, which implies $x = 1$, or equivalently, $N = U$.
Therefore, $h(0; N, U) \leq 1$ for all $0 < N \leq U$, with equality if and only if $N = U$. \hfill$\Box$

\begin{proposition}
Let $f_{\text{H}}(n; \eta_\text{init}, \eta_\text{inf}, N, U)$ be the learning rate function for HyperbolicLR where $\eta_\text{init} > \eta_\text{inf} > 0$ and $0 < N \leq U$. Then, at the maximum epoch $N$, we have:

\[ f_{\text{H}}(N; \eta_\text{init}, \eta_\text{inf}, N, U) \geq \eta_\text{inf} \]
\end{proposition}

\noindent \textit{Proof}
Recall that 
\begin{equation}
   f_\text{H}(n; \eta_\text{init}, \eta_\text{inf}, N, U) = \eta_\text{init} + (\eta_\text{init} - \eta_\text{inf}) (h(n; N, U) - h(0; N, U))
   \label{eq:hyperbolic_lr}
\end{equation}
At $n = N$, we have:
\begin{align*}
f_{\text{H}}(N; \eta_\text{init}, \eta_\text{inf}, N, U) &= \eta_\text{init} + (\eta_\text{init} - \eta_\text{inf})(0 - h(0; N, U)) \\
  &= \eta_\text{init} - (\eta_\text{init} - \eta_\text{inf}) h(0; N, U)
\end{align*}
From the previous proposition, we know that $h(0; N, U) \leq 1$ for all $0 < N \leq U$. Therefore:
\begin{align*}
f_{\text{H}}(N; \eta_\text{init}, \eta_\text{inf}, N, U) &= \eta_\text{init} - (\eta_\text{init} - \eta_\text{inf}) h(0; N, U) \\
  &\geq \eta_\text{init} - (\eta_\text{init} - \eta_\text{inf}) \\
  &= \eta_\text{inf}
\end{align*}
The equality holds when $N = U$, as in this case $h(0; N, U) = 1$.
Therefore, we conclude that $f_{\text{H}}(N; \eta_\text{init}, \eta_\text{inf}, N, U) \geq \eta_\text{inf}$ for all valid parameter values.\hfill$\Box$

\begin{proposition}
Let $f_{\text{EH}}(n; \eta_\text{init}, \eta_\text{inf}, N, U)$ be the learning rate function for ExpHyperbolicLR where $\eta_\text{init} > \eta_\text{inf} > 0$ and $0 < N \leq U$. Then, at the maximum epoch $N$, we have:

\[ f_{\text{EH}}(N; \eta_\text{init}, \eta_\text{inf}, N, U) \geq \eta_\text{inf} \]
\end{proposition}

\noindent \textit{Proof}
Recall that 
\begin{equation*}
f_{\text{EH}}(n; \eta_\text{init}, \eta_\text{inf}, N, U) = \exp(f_{\text{H}}(n; \ln \eta_\text{init}, \ln \eta_\text{inf}, N, U)),
\end{equation*}
where $f_{\text{H}}$ is the HyperbolicLR function.
From Proposition 3, we know that for HyperbolicLR:
\[ f_{\text{H}}(N; \ln \eta_\text{init}, \ln \eta_\text{inf}, N, U) \geq \ln \eta_\text{inf} \]
Applying this to the ExpHyperbolicLR function at $n = N$:
\begin{align*}
f_{\text{EH}}(N; \eta_\text{init}, \eta_\text{inf}, N, U) &= \exp(f_{\text{H}}(N; \ln \eta_\text{init}, \ln \eta_\text{inf}, N, U)) \\
  &\geq \exp(\ln \eta_\text{inf}) \\
  &= \eta_\text{inf}
\end{align*}
The equality holds when $N = U$ and $\eta_\text{init} = \eta_\text{inf}$.
Therefore, we conclude $f_{\text{EH}}(N; \eta_\text{init}, \eta_\text{inf}, N, U) \geq \eta_\text{inf}$ for all valid parameter values.\hfill$\Box$

\section{Dataset Generation Details}
\label{sec:dataset_generation}

\subsection{CIFAR-10 Augmentation}

For the CIFAR-10 dataset, we applied the following data augmentation techniques to the training set:
\begin{itemize}
    \item Random cropping to $32 \times 32$ pixels with 4-pixel padding
    \item Random horizontal flipping
    \item Per-channel standardization using mean values of (0.4914, 0.4822, 0.4465) and standard deviation values of (0.2023, 0.1994, 0.2010) for the respective channels
\end{itemize}

\subsection{Oscillation Dataset}

The oscillation dataset was generated using the following ordinary differential equation:

\begin{equation}
   m \ddot{u} + c \dot{u} + k u = 0
\end{equation}

where $m = 1$, $k = 200$, and $c = \zeta \cdot 2\sqrt{m k}$.
We varied the damping ratio $\zeta$ with values $0, 0.01$, and $0.02$ to simulate different oscillatory behaviors.
The ODE was numerically solved using the Newmark-beta method over the interval $t \in [0, 10]$ with a step size of $10^{-3}$.
Initial conditions were set as $u(0) = 0.1$, $\dot{u}(0) = 0$, and $\ddot{u}(0) = -20$.

We employed a sliding window approach to prepare the data for the prediction task.
For each $\zeta$ value, we generated 10,001 data points and applied the sliding window technique with a history of 100 time steps and a prediction horizon of 20 time steps.
The resulting sequences from all three $\zeta$ values were then combined into a single dataset.

This process yielded a total of 29,646 input-output pairs, with each input sequence comprising 100 consecutive time steps used to predict the subsequent 20 time steps.
The final dataset had an input size of $(29646, 100, 1)$ and a label size of $(29646, 20, 1)$.

To ensure a random distribution of oscillation patterns across different damping ratios, we shuffled the combined dataset before splitting it into training and validation sets.
We used an 80-20 split, resulting in 23,716 samples for training and 5,930 samples for validation.
Prior to training, we normalized the data to the range $[0, 1]$ to ensure consistent scale across different oscillation parameters.

\subsection{Integral Dataset}

For the operator learning task, we aimed to learn an operator $G$ defined for continuous functions $u \in C[0,1]$ and arbitrary real numbers $y \in [0,1]$ as follows:

\begin{equation}
   G(u)(y) = \int_0^y u(x) \mathrm{d}x
\end{equation}

To accomplish this, we employed the DeepONet approach \cite{lu2021learning}.
Training a DeepONet requires two distinct inputs: values of the input function $u$ at fixed sensor points $x_i$, denoted as $u(x_i)$, and domain values $y_i$ of the operator-mapped function $G(u)$.
The corresponding labels are the values of $G(u)(y_i)$.

For the target points $y_i$, we uniformly partitioned the interval $[0,1]$ into 100 points.
To generate inputs corresponding to the input functions, we employed a Gaussian Random Field approach with a squared exponential kernel.
To enhance the diversity of our dataset, we introduced variability in the length scale parameter $l$, uniformly sampling it from the range $[0.1, 0.4]$.

For each randomly selected length scale parameter $l$, we uniformly divided the $[0,1]$ interval into 100 sensor points.
We then generated a Gaussian Random Field $X_i$ corresponding to each point $x_i$, considering this as a discrete representation $\mathbf{u}$ of a continuous function $u$:

\begin{equation}
   \begin{aligned}
      \mathbf{u}  &= [X_0, X_1, \ldots, X_{98}, X_{99}] \\
                  &= [u(0), u(x_1), \ldots, u(x_{98}), u(1)]
   \end{aligned}
\end{equation}

We generated 10,000 such functions.
To match the size, we replicated the vector $\mathbf{y}$ of target points $y_i$ 10,000 times.
Consequently, the input sizes are $(10000, 100)$ for $\mathbf{u}$ and $(10000, 100)$ for $\mathbf{y}$.

We employed a Gaussian Random Field approach with a squared exponential kernel to generate input functions.
The length scale parameter $l$ was uniformly sampled from the range $[0.1, 0.4]$ to enhance dataset diversity.

\section{Model Architectures}
\label{sec:model_architectures}

This section provides detailed descriptions of the model architectures used in our experiments.

\begin{figure*}[t]
   \centering
   \includegraphics[width=.95\linewidth]{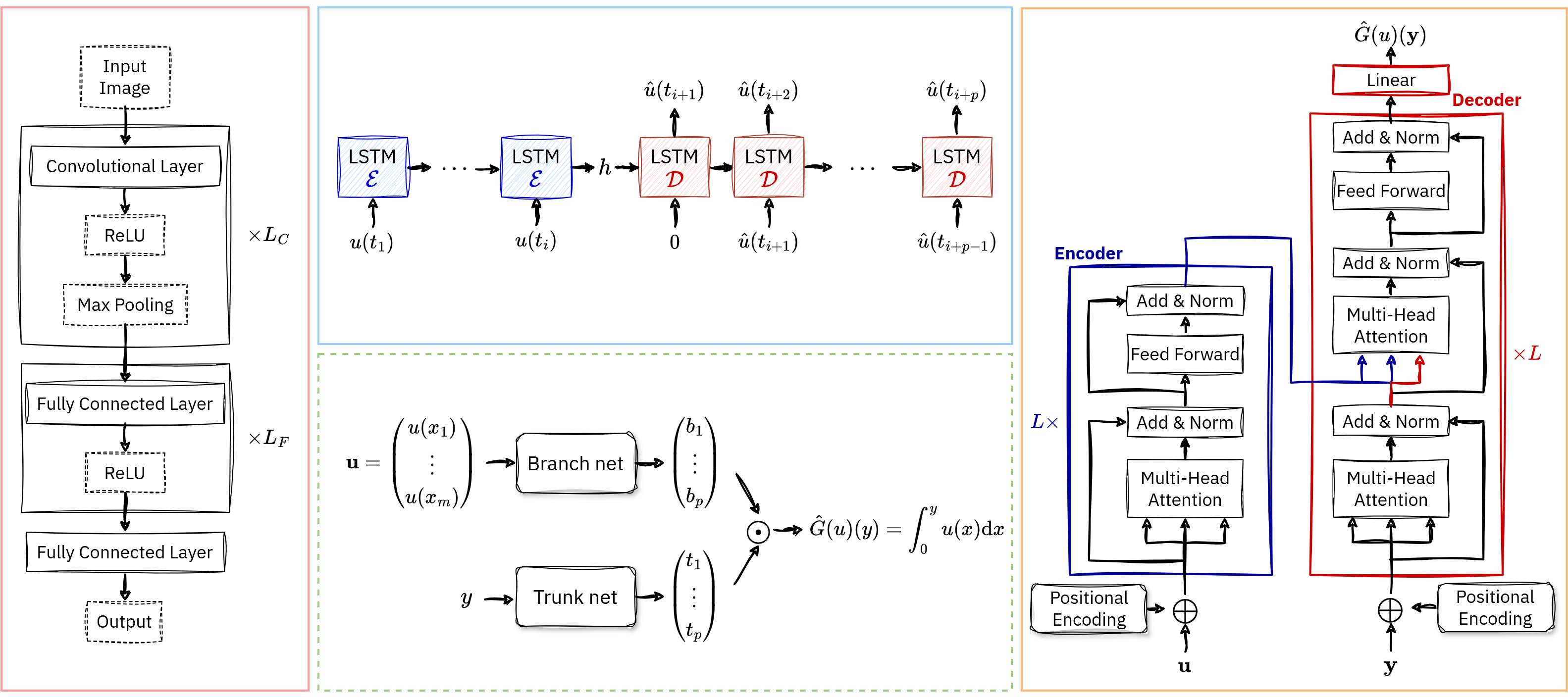}
   \caption{
    \textcolor{Maroon}{(Red solid box)} Architectures of SimpleCNN for CIFAR-10 classification, \textcolor{blue}{(Blue solid box)} LSTM Sequence-to-Sequence model for oscillation prediction,
    \textcolor{Olive}{(Green dashed box)} DeepONet model for learning integral operator and \textcolor{orange}{(Orange solid box)} TraONet model for learning integral operator.
   }
   \label{fig:model_architectures}
\end{figure*}

\subsection{SimpleCNN for CIFAR-10}

As shown in the red solid box of Figure \ref{fig:model_architectures}, our SimpleCNN architecture for CIFAR-10 classification consists of a series of convolutional layers followed by fully connected layers.
Each convolutional block, repeated $L_C$ times, comprises a convolutional layer, ReLU activation, and max pooling.
This structure allows for hierarchical feature extraction from the input images.
Following the convolutional blocks, the network includes $L_F + 1$ fully connected layers.
The first $L_F$ layers are followed by ReLU activation, while the final layer produces the output logits for classification.

\subsection{LSTM Seq2Seq for oscillation prediction}

The LSTM Sequence-to-Sequence model, depicted in the blue solid box of Figure \ref{fig:model_architectures}, consists of an encoder ($\mathcal{E}$) and a decoder ($\mathcal{D}$), both utilizing LSTM layers.
The encoder processes the input sequence $u(t_1), u(t_2), ..., u(t_i)$, representing historical oscillation data.
The final hidden state $h$ of the encoder is used to initialize the decoder.
The decoder operates autoregressively, generating predictions $\hat{u}(t_{i+1}), \hat{u}(t_{i+2}), \ldots, \hat{u}(t_{i+p})$ for future time steps.
It takes its own previous output as input for the next prediction, starting with an initial input of 0.
This design challenges the model to maintain long-term coherence in its predictions, crucial for accurately forecasting oscillatory behavior.

\subsection{DeepONet for operator learning}

As illustrated in the green dashed box of Figure \ref{fig:model_architectures}, we employed the DeepONet architecture for our operator learning task.
Both the branch and trunk networks consist of $L$ repeated layers of MLP followed by GELU activation, with a final MLP layer without activation.
The outputs of these networks, $\mathbf{b} = [b_1, b_2, ..., b_p]$ from the branch net and $\mathbf{t} = [t_1, t_2, ..., t_p]$ from the trunk net (where $p$ is a hyperparameter), are combined through an inner product to yield the operator output $\hat{G}(u)(y)$.

\subsection{TraONet for operator learning}

To enhance the efficiency of operator learning, we developed and utilized \textit{TraONet} (Transformer Operator Network), a novel operator network that incorporates Transformer encoder and decoder models alongside the traditional DeepONet structure.
As shown in the orange solid box of Figure \ref{fig:model_architectures}, TraONet draws inspiration from the Transformer architecture introduced in the original Transformer paper \cite{vaswani2017attention}.
However, we adapted the structure for operator learning by removing the masking in the decoder's masked multi-head attention component.
Both the encoder and decoder parts of TraONet are repeated $L$ times to achieve the desired depth of representation.
This novel approach combines the strengths of attention mechanisms with the proven effectiveness of DeepONet, potentially offering improved performance in capturing complex operator relationships.

\section{Experimental Design and Evaluation Protocol}
\label{sec:experimental_design}

\subsection{Model hyperparameter optimization}

The hyperparameters of our models primarily consist of discrete values, such as the number of layers and nodes.
Consequently, we employed a grid search within predefined ranges to optimize these hyperparameters.
For consistency across tasks, we utilized the ExponentialLR scheduler, which has the fewest hyperparameters, fixing the decay factor at 0.9 for all experiments over 50 epochs.
The initial learning rate was task-specific: $10^{-2}$ for CIFAR-10 and the oscillation dataset, and $5 \times 10^{-3}$ for the integral dataset.
Table \ref{tab:grid_search_ranges} presents the grid search ranges for each model architecture along with the optimal values determined through our comprehensive search.

\begin{table}[ht]
\caption{Grid search ranges for hyperparameter optimization of each model architecture.}
\label{tab:grid_search_ranges}
\vskip 0.1in
\begin{center}
\begin{small}
\begin{sc}
\begin{tabular}{cccc}
\toprule
\textbf{Model} & \textbf{Parameter name} & \textbf{Candidates} & \textbf{Optimal value} \\
\midrule
\multirow{4}{*}{SimpleCNN} & Convolutional layers ($L_C$) & $\{2,3,4,5\}$ & $4$ \\
& Fully connected layers ($L_F$) & $\{2,3,4,5\}$ & $2$ \\
& Convolutional channels & $\{32,64,128,256\}$ & $128$ \\
& Fully connected units & $\{128,256,512,1024\}$ & $256$ \\
\midrule
\multirow{2}{*}{LSTM Seq2Seq} & Hidden sizes & $\{64,128,256,512\}$ & $128$ \\
& Hidden layers & $\{2,3,4,5\}$ & $4$ \\
\midrule
\multirow{3}{*}{DeepONet} & Hidden sizes & $\{64,128,256,512,1024\}$ & $256$ \\
& Hidden layers & $\{3,4,5,6\}$ & $4$ \\
& Branches ($p$) & $\{10,20,30,40\}$ & $10$ \\
\midrule
\multirow{4}{*}{TraONet} & Model dimension & $\{8,16,32,64,128\}$ & $32$ \\
& Attention heads & $\{2,4,8\}$ & $2$ \\
& Feed-forward dimension & $\{64,128,256,512\}$ & $128$ \\
& Layers ($L$) & $\{2,3,4\}$ & $2$ \\
\bottomrule
\end{tabular}
\end{sc}
\end{small}
\end{center}
\vskip -0.1in
\end{table}

These optimized configurations form the foundation for our subsequent experiments, ensuring that each model architecture is well-tuned for its respective task before proceeding to scheduler optimization.
The grid search ranges were carefully selected to cover a wide spectrum of model complexities while remaining computationally feasible.
The optimal values obtained from this process represent a balance between model capacity and computational efficiency, tailored to the specific requirements of each task and dataset.

\subsection{Scheduler hyperparameter optimization}

Following the optimization of model hyperparameters, we focused on optimizing the scheduler hyperparameters.
Unlike the discrete model parameters, scheduler hyperparameters are primarily continuous values (e.g., learning rate, power, decay factor, infimum learning rate and etc.).
To effectively optimize these continuous parameters, we employed the \textit{Tree-structured Parzen Estimator} (TPE), a variant of Bayesian optimization \cite{bergstra2011algorithms,watanabe2023tree}.

Our primary objective was to assess whether optimizations performed at lower epoch counts could consistently maintain performance as the number of epochs increased.
Therefore, we fixed the number of epochs at 50 for this optimization phase.
For each task and scheduler combination, we conducted 25 trials using TPE, selecting the optimal values from these trials.
\begin{table}[h]
\caption{Hyperparameter optimization ranges for each scheduler.}
\label{tab:scheduler_ranges}
\vskip 0.1in
\begin{center}
\begin{small}
\begin{sc}
\begin{tabular}{ccc}
\toprule
\textbf{Scheduler} & \textbf{Parameter} & \textbf{Optimization Range} \\
\midrule
PolynomialLR & $p$ & 0.5 - 3.0 \\ 
\midrule
CosineAnnealingLR & $\eta_\text{min}$ & $10^{-7}$ - $10^{-4}$ (log scale) \\
\midrule
ExponentialLR & $\gamma$ & 0.9 - 0.99 \\
\midrule
HyperbolicLR \& & $\eta_\text{inf}$ & $10^{-7}$ - $10^{-4}$ (log scale) \\
ExpHyperbolicLR & $U$ & 200 - 400 (step = 50) \\
\bottomrule
\end{tabular}
\end{sc}
\end{small}
\end{center}
\end{table}
The initial learning rate optimization range was consistent across all schedulers, spanning from $10^{-4}$ to $5 \times 10^{-2}$ on a logarithmic scale.
Other hyperparameters were optimized within scheduler-specific ranges.
The detailed optimization ranges for each scheduler's hyperparameters are presented in Table \ref{tab:scheduler_ranges}.
\begin{table}[h]
\caption{Optimal hyperparameters for each scheduler and model combination. Scheduler abbreviations: N = No scheduler (constant learning rate), P = PolynomialLR, C = CosineAnnealingLR, E = ExponentialLR, H = HyperbolicLR, EH = ExpHyperbolicLR.}
\label{tab:optimal_hyperparameters}
\begin{center}
\begin{small}
\begin{sc}
\begin{tabular}{ccccccc}
\toprule
\textbf{Scheduler} & \textbf{Parameter} & \textbf{SimpleCNN} & \textbf{LSTM Seq2Seq} & \textbf{DeepONet} & \textbf{TraONet} \\
\midrule
N & $\eta_\text{init}$ & $6.15 \times 10^{-4}$ & $1.05 \times 10^{-4}$ & $1.03 \times 10^{-3}$ & $7.27 \times 10^{-4}$ \\
\midrule
\multirow{2}{*}{P} & $\eta_\text{init}$ & $7.92 \times 10^{-4}$ & $3.96 \times 10^{-3}$ & $4.13 \times 10^{-3}$ & $7.36 \times 10^{-4}$ \\
& $p$ & $0.7609$ & $1.2752$ & $1.2443$ & $0.5319$ \\
\midrule
\multirow{2}{*}{C} & $\eta_\text{init}$ & $1.06 \times 10^{-3}$ & $3.00 \times 10^{-3}$ & $4.62 \times 10^{-3}$ & $1.59 \times 10^{-3}$ \\
& $\eta_\text{min}$ & $2.13 \times 10^{-5}$ & $1.10 \times 10^{-7}$ & $2.66 \times 10^{-7}$ & $2.61 \times 10^{-6}$ \\
\midrule
\multirow{2}{*}{E} & $\eta_\text{init}$ & $5.91 \times 10^{-4}$ & $2.68 \times 10^{-3}$ & $2.01 \times 10^{-3}$ & $1.53 \times 10^{-3}$ \\
& $\gamma$ & $0.9894$ & $0.9392$ & $0.9598$ & $0.9649$ \\
\midrule
\multirow{3}{*}{H} & $\eta_\text{init}$ & $9.39 \times 10^{-4}$ & $2.44 \times 10^{-3}$ & $4.62 \times 10^{-3}$ & $2.55 \times 10^{-3}$ \\
& $\eta_\text{inf}$ & $9.47 \times 10^{-6}$ & $5.99 \times 10^{-6}$ & $2.66 \times 10^{-7}$ & $1.69 \times 10^{-5}$ \\
& $U$ & $400$ & $200$ & $250$ & $250$ \\
\midrule
\multirow{3}{*}{EH} & $\eta_\text{init}$ & $9.50 \times 10^{-4}$ & $2.44 \times 10^{-3}$ & $4.59 \times 10^{-3}$ & $1.03 \times 10^{-3}$ \\
& $\eta_\text{inf}$ & $5.40 \times 10^{-5}$ & $5.99 \times 10^{-6}$ & $5.74 \times 10^{-7}$ & $7.11 \times 10^{-5}$ \\
& $U$ & $350$ & $200$ & $250$ & $350$ \\
\bottomrule
\end{tabular}
\end{sc}
\end{small}
\end{center}
\end{table}
The optimal hyperparameters for each scheduler and model combination are summarized in Table \ref{tab:optimal_hyperparameters}.
These values represent the best performing configurations within our specified ranges, as determined by the TPE optimization process.

These optimized scheduler configurations form the basis for our subsequent experiments, where we evaluate the performance of each scheduler across varying epoch settings.
By optimizing at a fixed, lower epoch count (50), we aim to investigate whether these configurations can maintain consistent performance improvements as we increase the number of epochs in our final evaluation phase.

\subsection{Performance evaluation}

To assess the efficacy of our optimized scheduler configurations across varying training durations, we conducted a series of experiments using the optimal hyperparameters presented in Table \ref{tab:optimal_hyperparameters}.
For each task, we progressively increased the number of epochs from 50 to 200 in increments of 50, measuring the validation loss for all models and accuracy for the SimpleCNN model on the CIFAR-10 task.
We evaluated the relative performance improvement per 50-epoch interval, calculating the mean and standard deviation to determine which scheduler demonstrated the most consistent and substantial performance gains.

Recognizing that the relationship between epochs and validation metrics (loss or accuracy) is not necessarily linear, we performed regression analysis to model this relationship more accurately.
After comparing linear, logarithmic, and power regression models, we found that power regression consistently yielded the highest significance across all tasks.
Consequently, we utilized power regression to compute $R^2$, exponent, and $p$-value for each scheduler-model combination.
These metrics provide insights into the potential for continued improvement and the consistency of the learning trajectory.

It's worth noting that for the CIFAR-10 task, all models except ExpHyperbolicLR exhibited $p$-values greater than 0.05 when regressing against validation loss.
However, when regressing against accuracy, all models showed $p$-values below 0.05.
This discrepancy may be attributed to the use of label smoothing (factor 0.1) during training.
While label smoothing can affect loss function behavior, potentially obscuring trends in validation loss, accuracy measurements remain less sensitive to these effects, explaining the more consistent statistical significance in accuracy-based regressions.

To quantify the extent of learning curve decoupling, we employed the smoothed learning curve difference metric introduced in Equation 6 in the main paper.
We applied this metric to all learning curves, using a Savitzky-Golay filter with a window size of 9 as our smoothing operator $\mathcal{S}$.

This comprehensive evaluation framework allows us to assess not only the raw performance of each scheduler but also the consistency of improvement and the degree to which the learning curves remain coupled across different training durations.
By combining these various metrics, we aim to provide a granular understanding of each scheduler's behavior and efficacy across our diverse set of deep learning tasks.

\subsection{Technical details}

Custom data for this study was generated using Rust-based libraries.
The oscillation dataset was created using Peroxide \cite{peroxide}, a numerical library, while the integral dataset utilized both Peroxide and RugField \cite{rugfield}, a Gaussian Random Field library.
Our deep learning experiments were conducted using Python 3.12.4 and PyTorch 2.3.1 \cite{paszke2017automatic}.
Data preprocessing was performed with Polars 1.1.0 \cite{polars} and NumPy 1.26.4 \cite{harris2020array}.
We employed Weights \& Biases \cite{wandb} for experiment logging and visualization.
Hyperparameter optimization was carried out using Optuna \cite{optuna_2019}, implementing the Tree-structured Parzen Estimator (TPE) algorithm.
All visualizations presented in this study were generated using Matplotlib \cite{Hunter:2007}, enhanced with the SciencePlots \cite{SciencePlots} package to ensure consistency with scientific publication standards.
The deep learning models were trained on one NVIDIA GeForce RTX 3080 GPU.

\section{Learning Curves}
\label{sec:learning_curves}

\begin{figure}[h]
\captionsetup[subfloat]{labelformat=empty}  %
\vskip 0.1in
\begin{center}
\subfloat[]{
    \includegraphics[width=.47\textwidth]{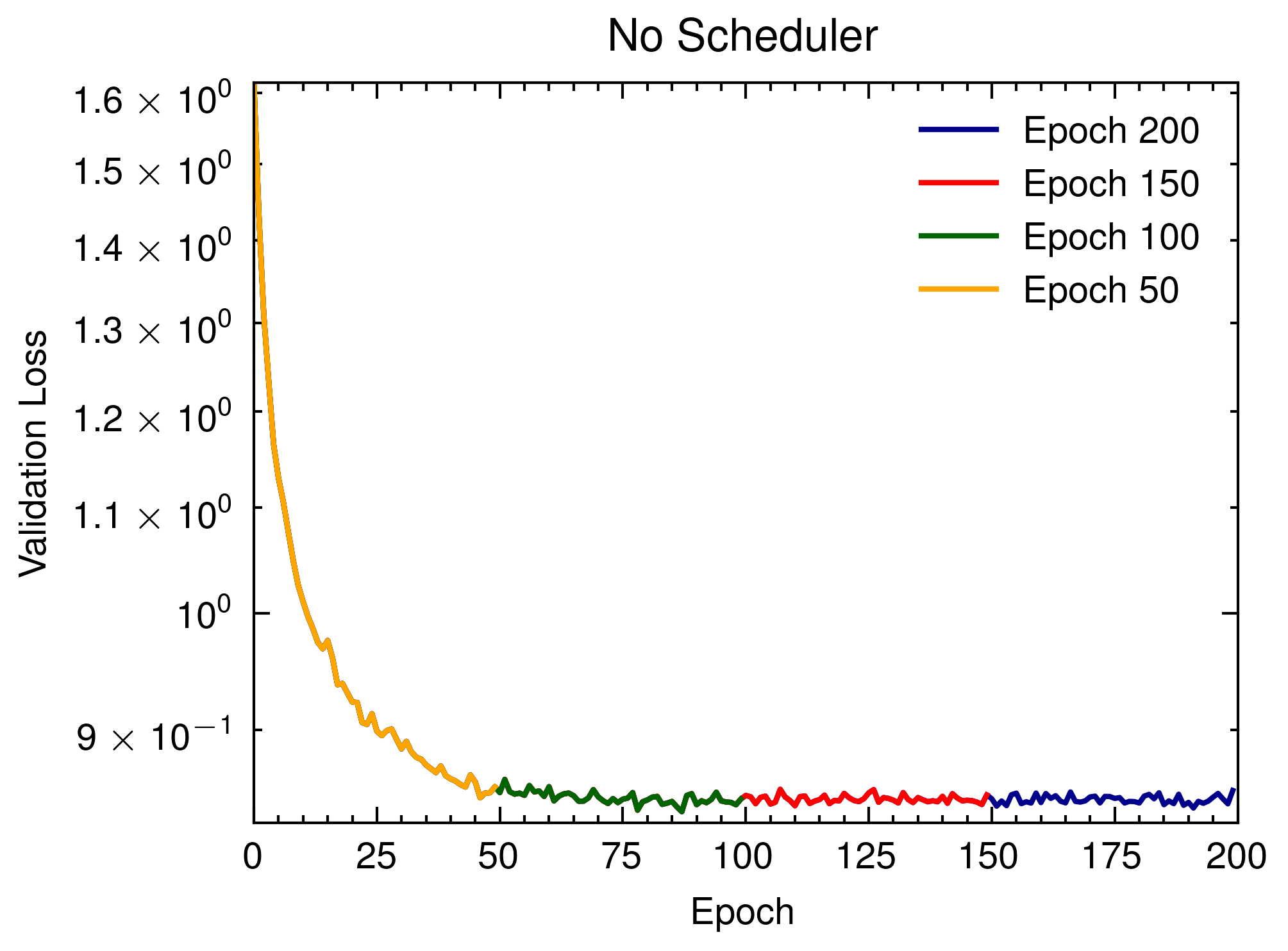}
}
\subfloat[]{
    \includegraphics[width=.47\textwidth]{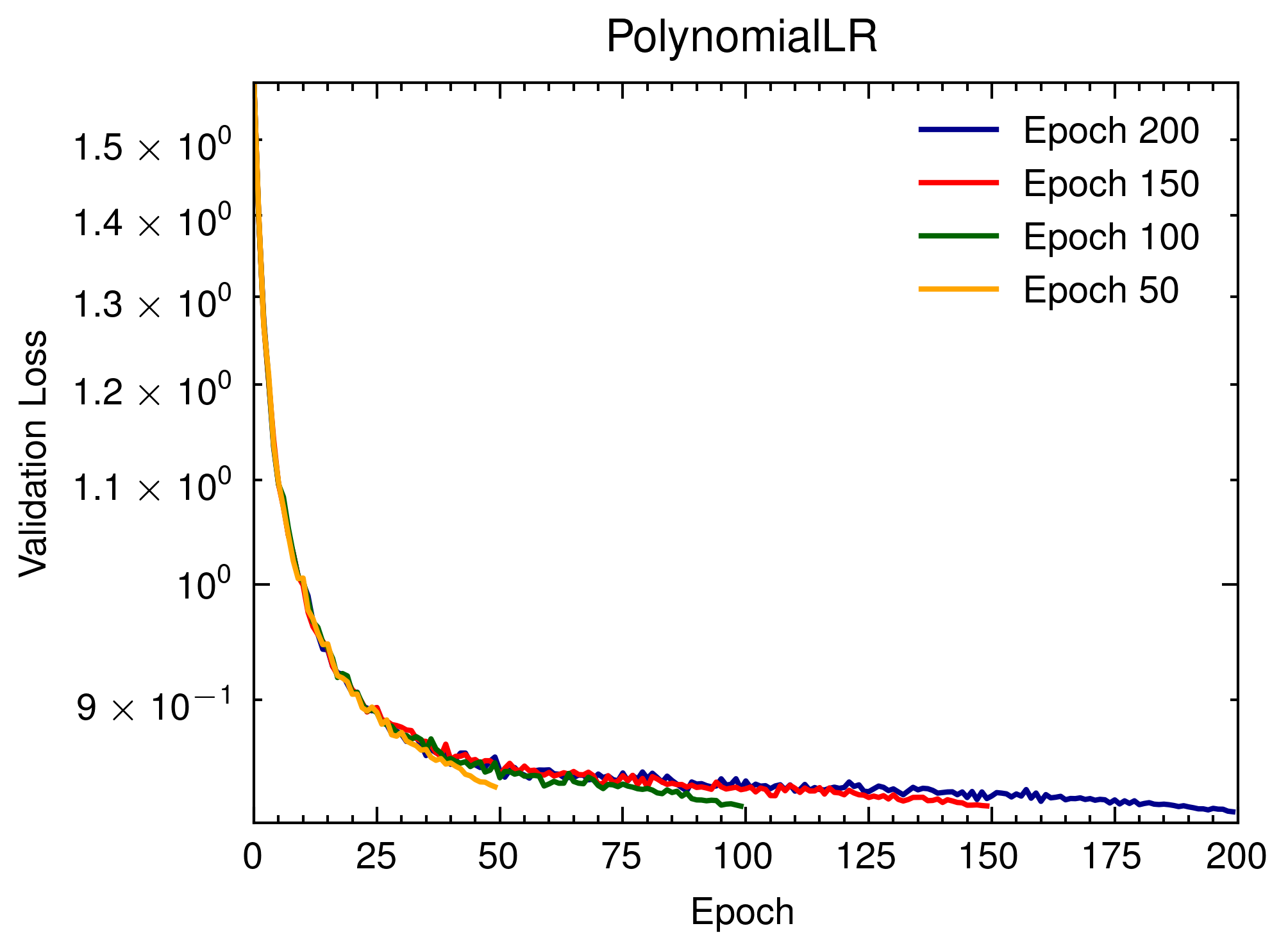}
}
\\
\subfloat[]{
    \includegraphics[width=.47\textwidth]{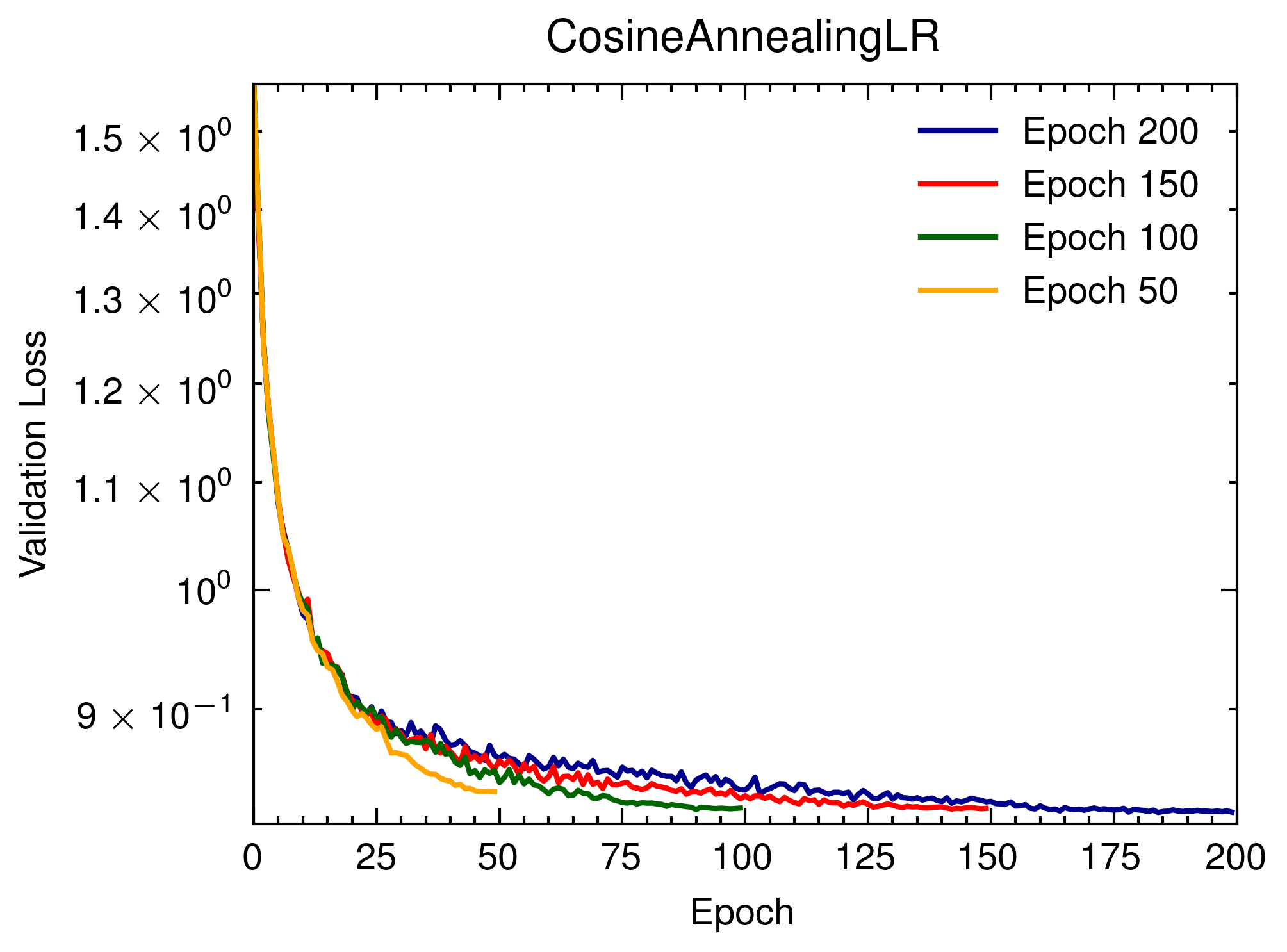}
}
\subfloat[]{
    \includegraphics[width=.47\textwidth]{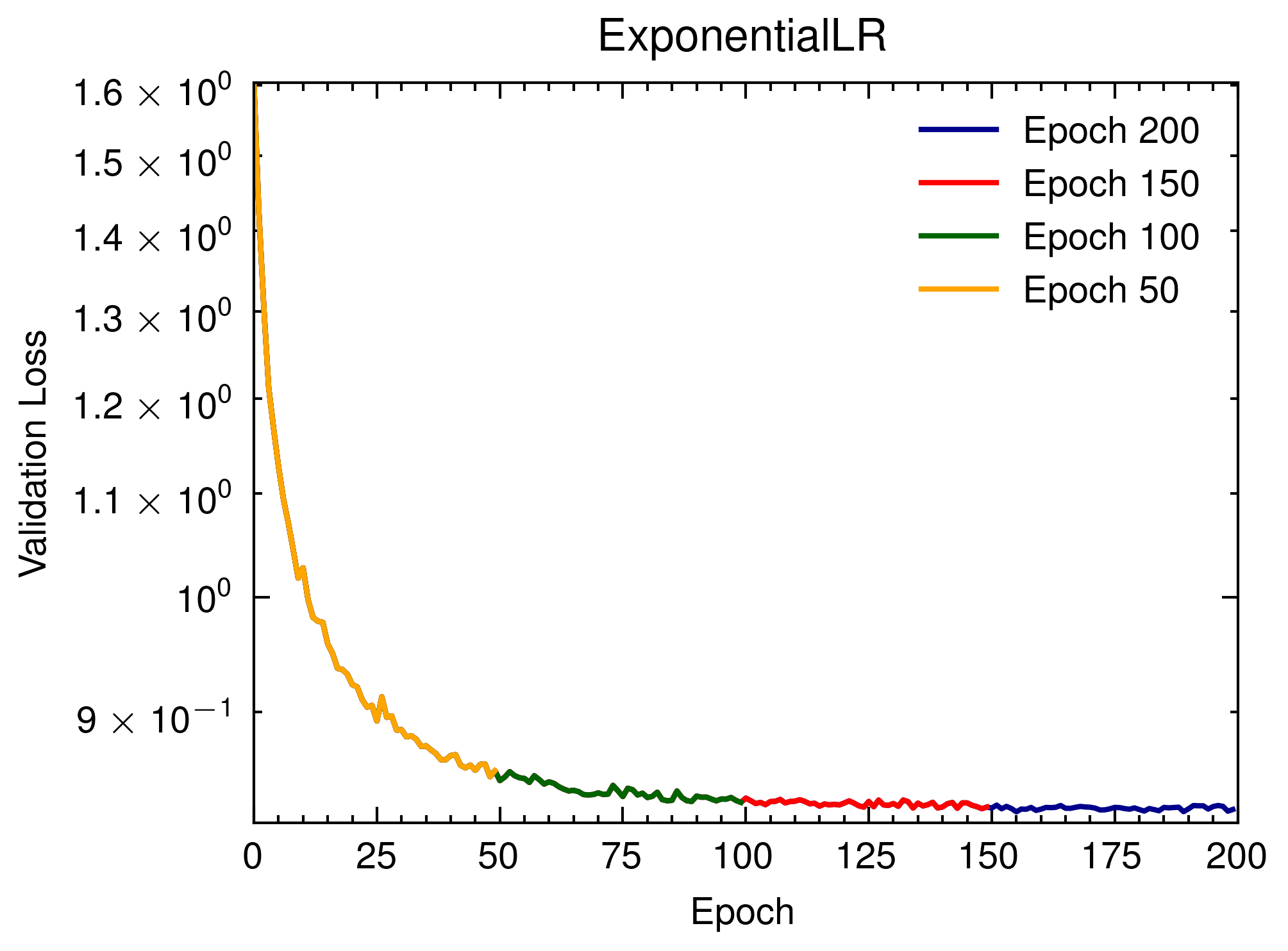}
}
\\
\subfloat[]{
    \includegraphics[width=.47\textwidth]{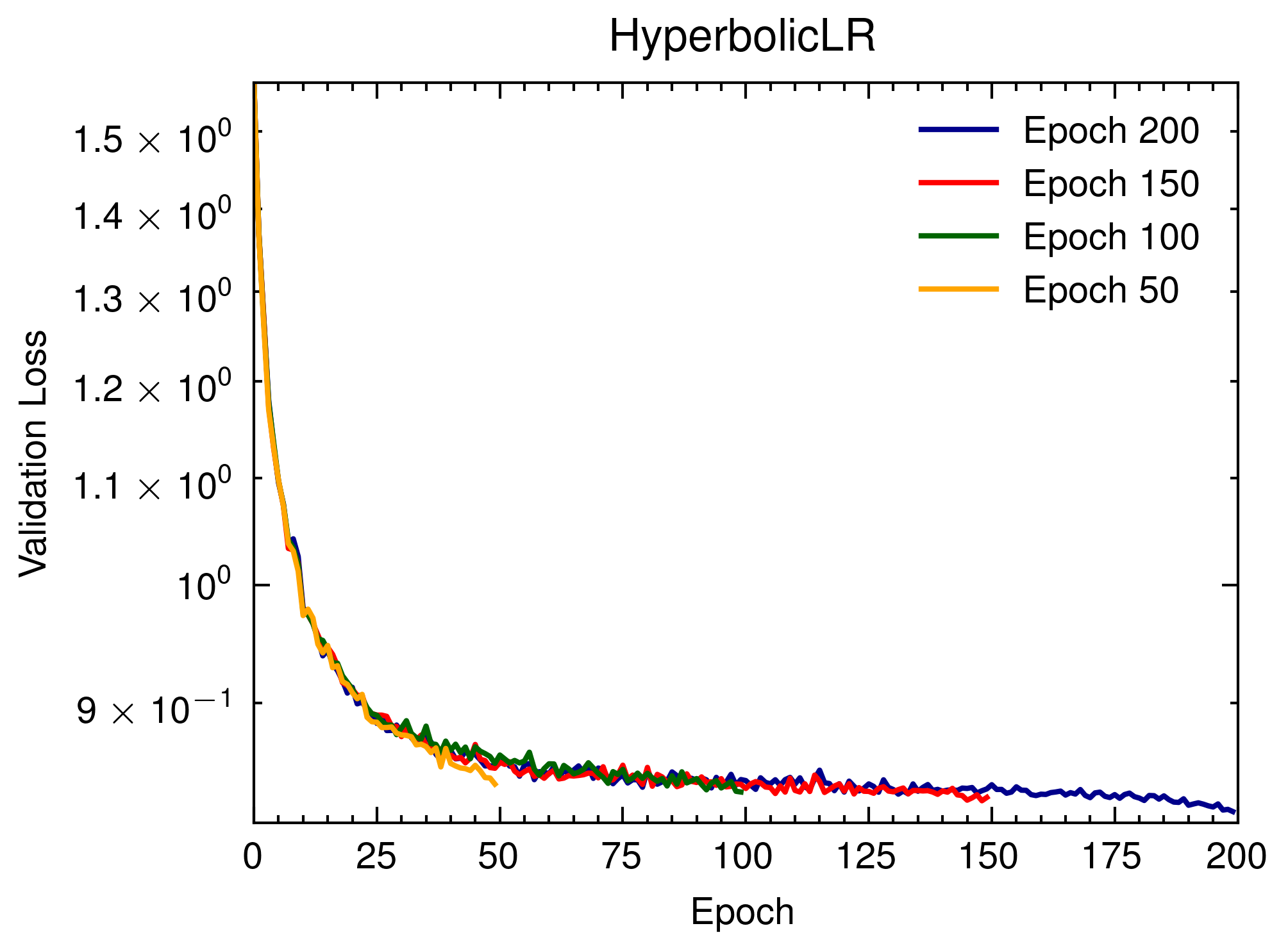}
}
\subfloat[]{
    \includegraphics[width=.47\textwidth]{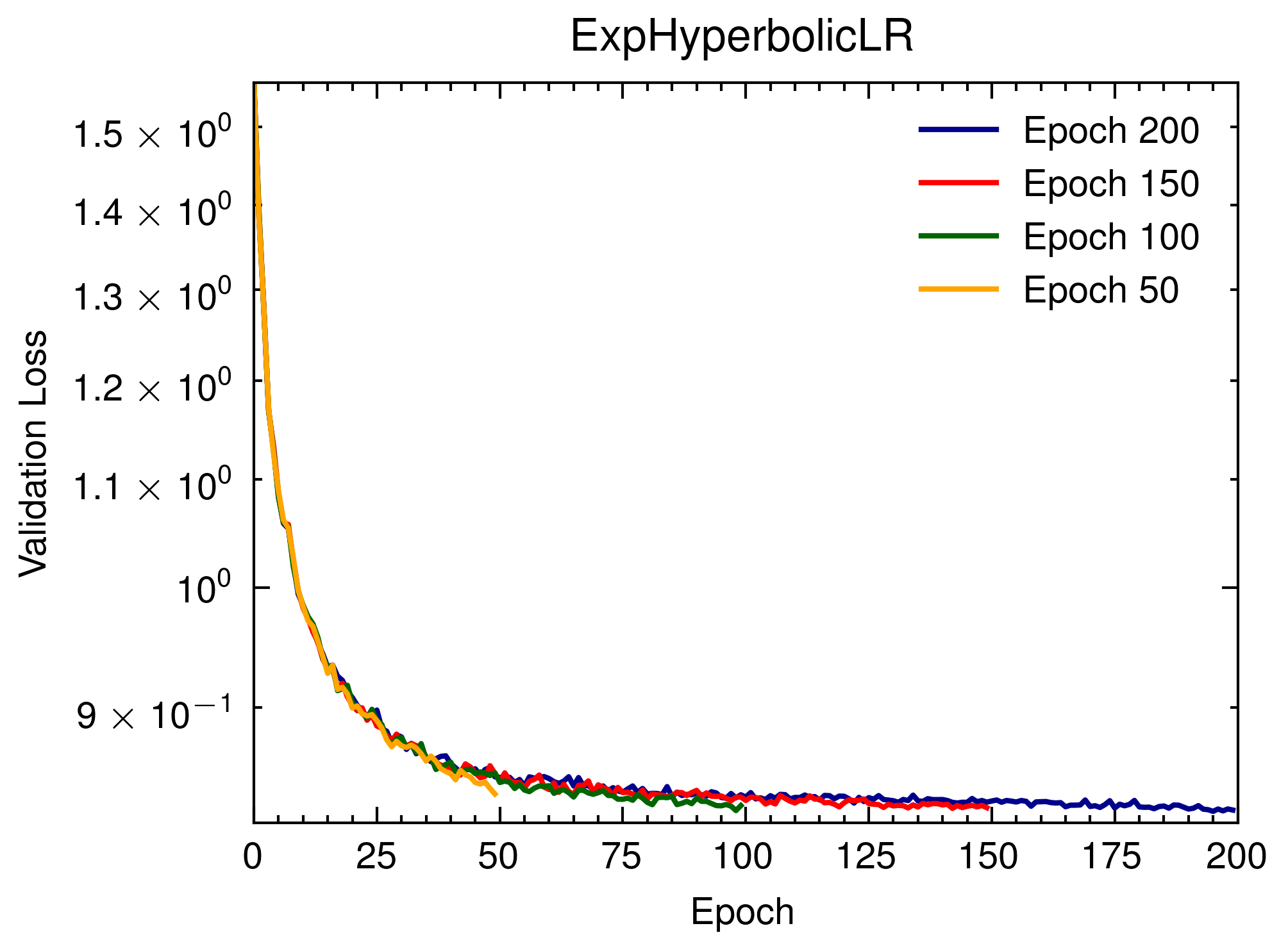}
}
\caption{Learning curves of validation loss for SimpleCNN on CIFAR-10.}
\label{fig:learning_curves_CIFAR10_CNN}
\end{center}
\vskip -0.1in
\end{figure}

\begin{figure}[h]
\captionsetup[subfloat]{labelformat=empty}
\vskip 0.1in
\begin{center}
\subfloat[]{
    \includegraphics[width=.47\textwidth]{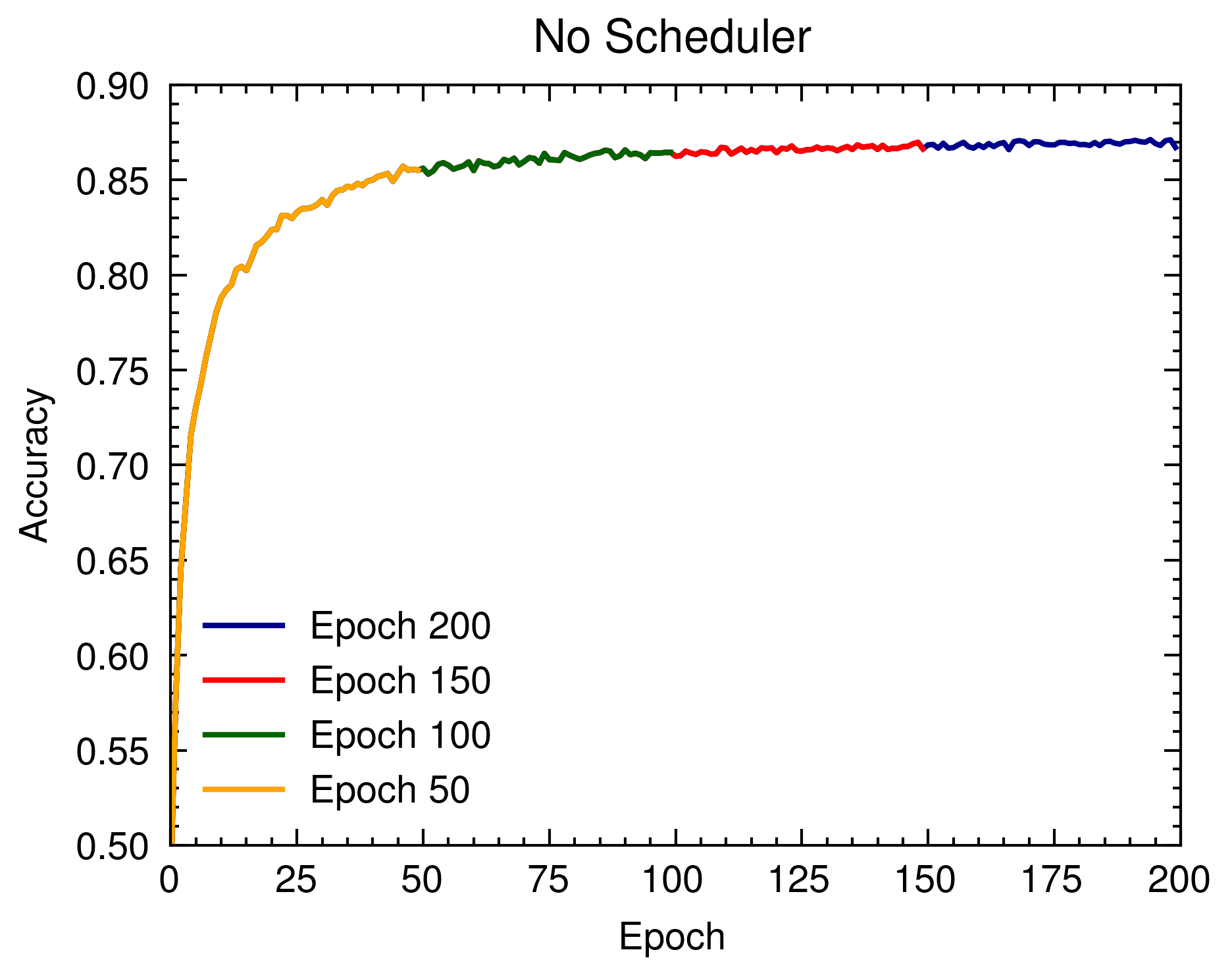}
}
\subfloat[]{
    \includegraphics[width=.47\textwidth]{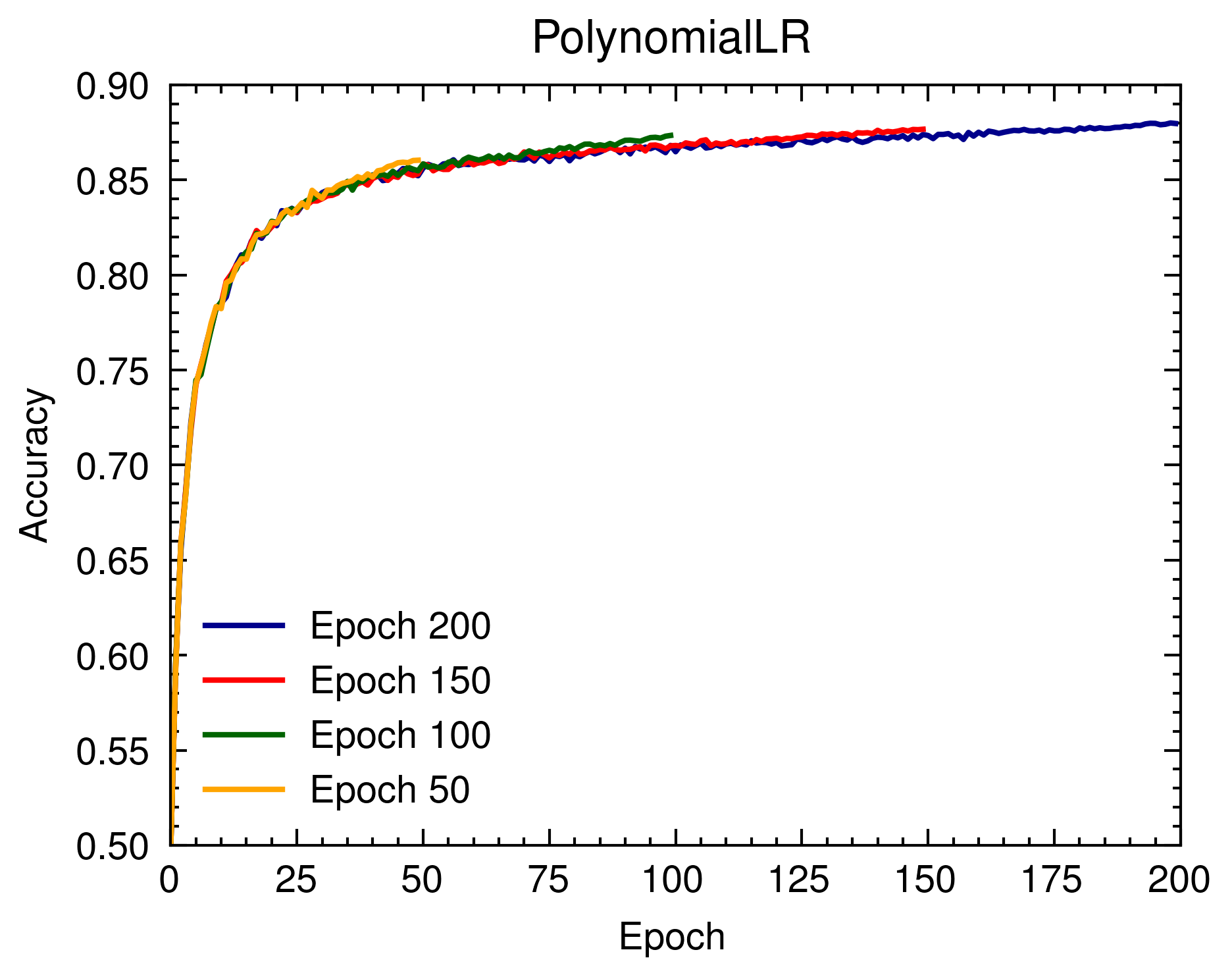}
}
\\
\subfloat[]{
    \includegraphics[width=.47\textwidth]{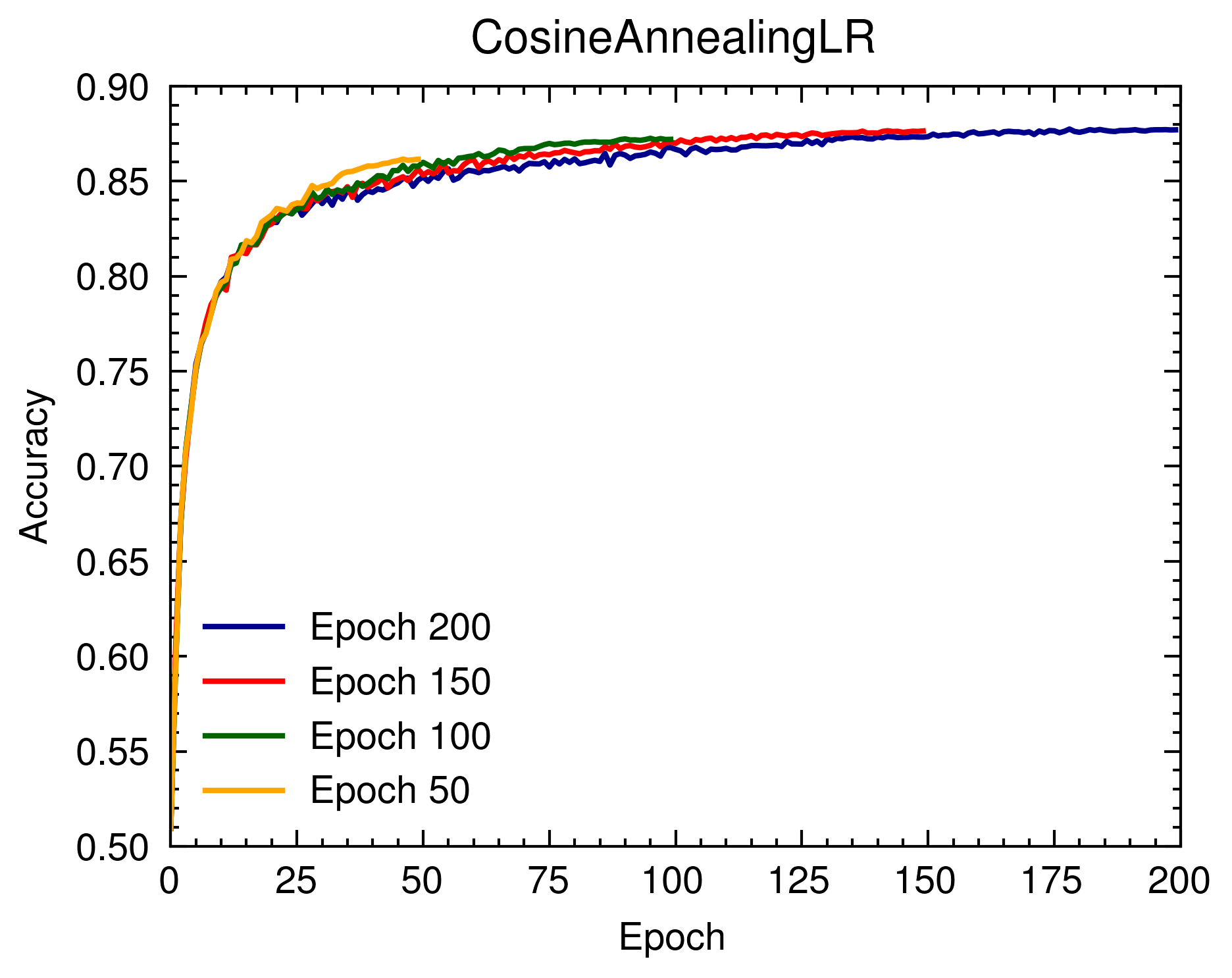}
}
\subfloat[]{
    \includegraphics[width=.47\textwidth]{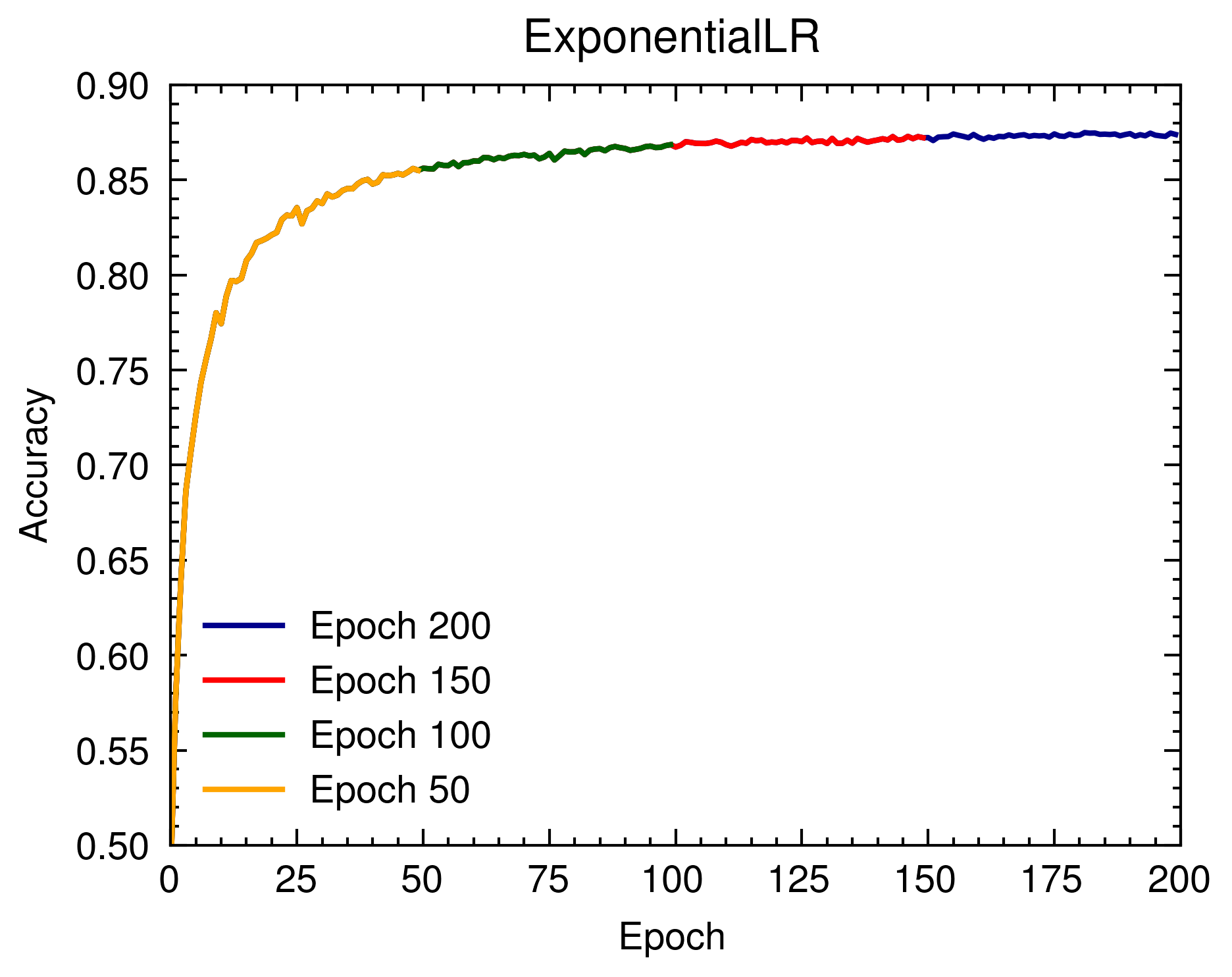}
}
\\
\subfloat[]{
    \includegraphics[width=.47\textwidth]{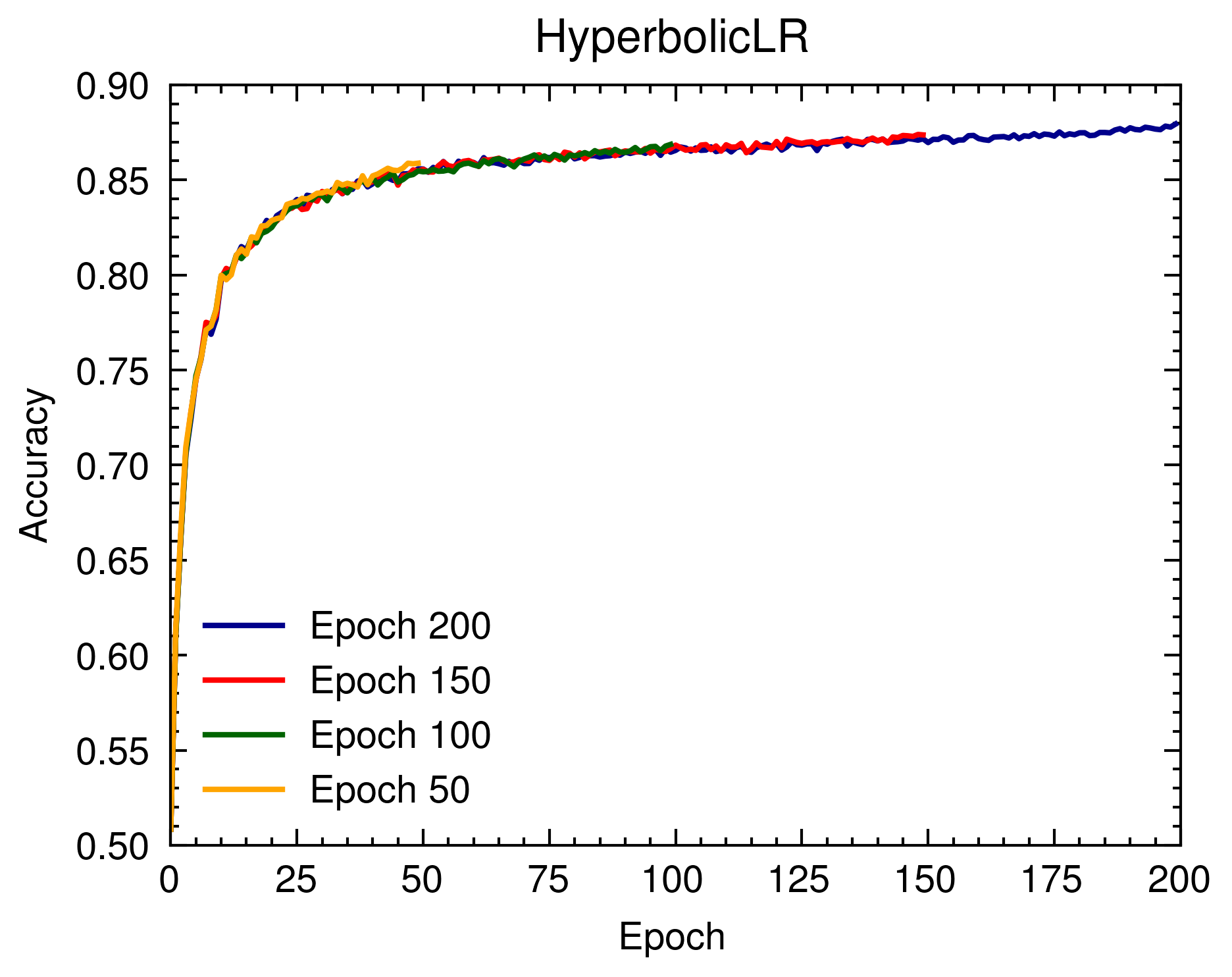}
}
\subfloat[]{
    \includegraphics[width=.47\textwidth]{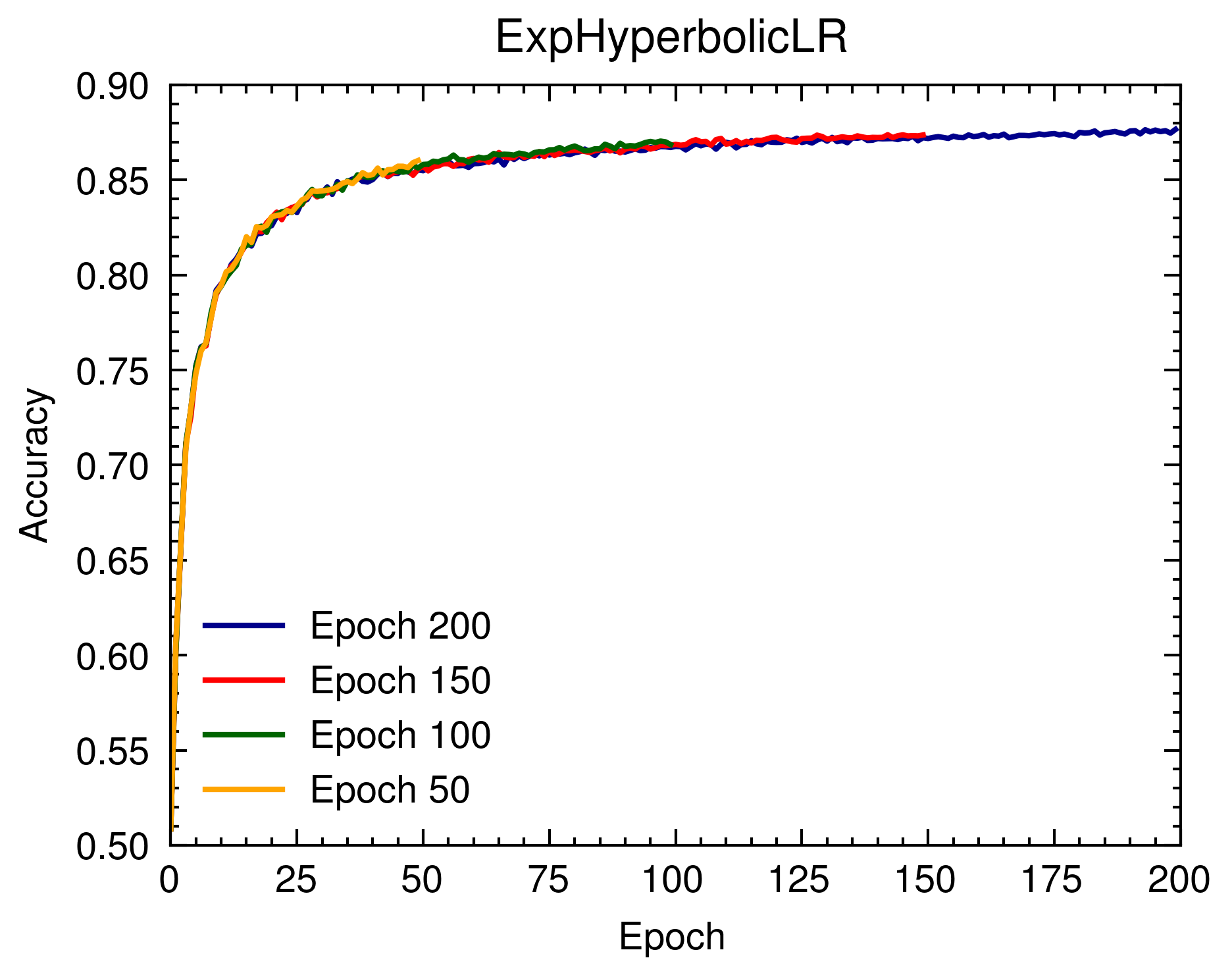}
}
\caption{Learning curves of validation accuracy for SimpleCNN on CIFAR-10.}
\label{fig:learning_curves_CIFAR10_CNN_ACC}
\end{center}
\end{figure}

\begin{figure}[h]
\captionsetup[subfloat]{labelformat=empty}
\vskip 0.1in
\begin{center}
\subfloat[]{
    \includegraphics[width=.47\textwidth]{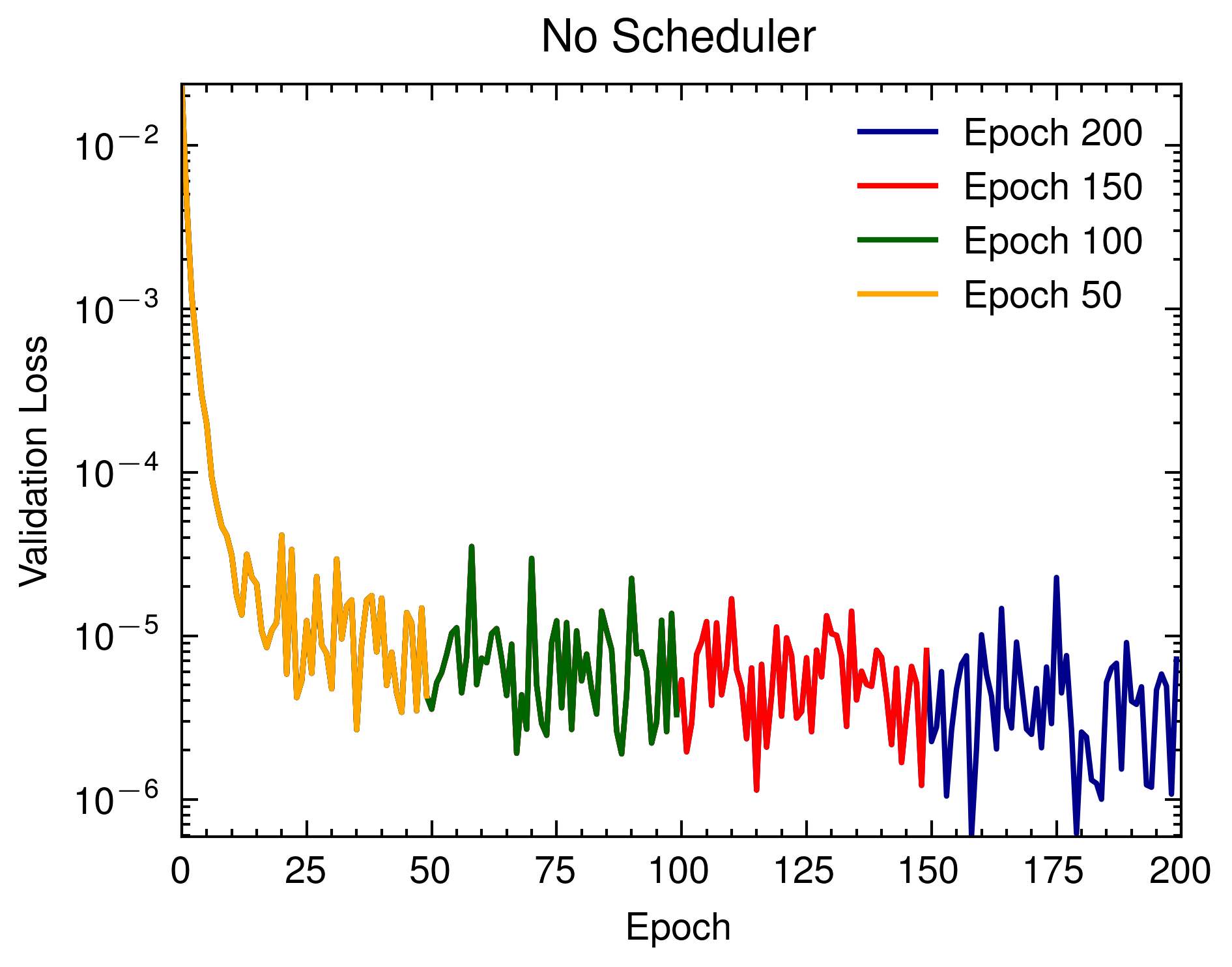}
}
\subfloat[]{
    \includegraphics[width=.47\textwidth]{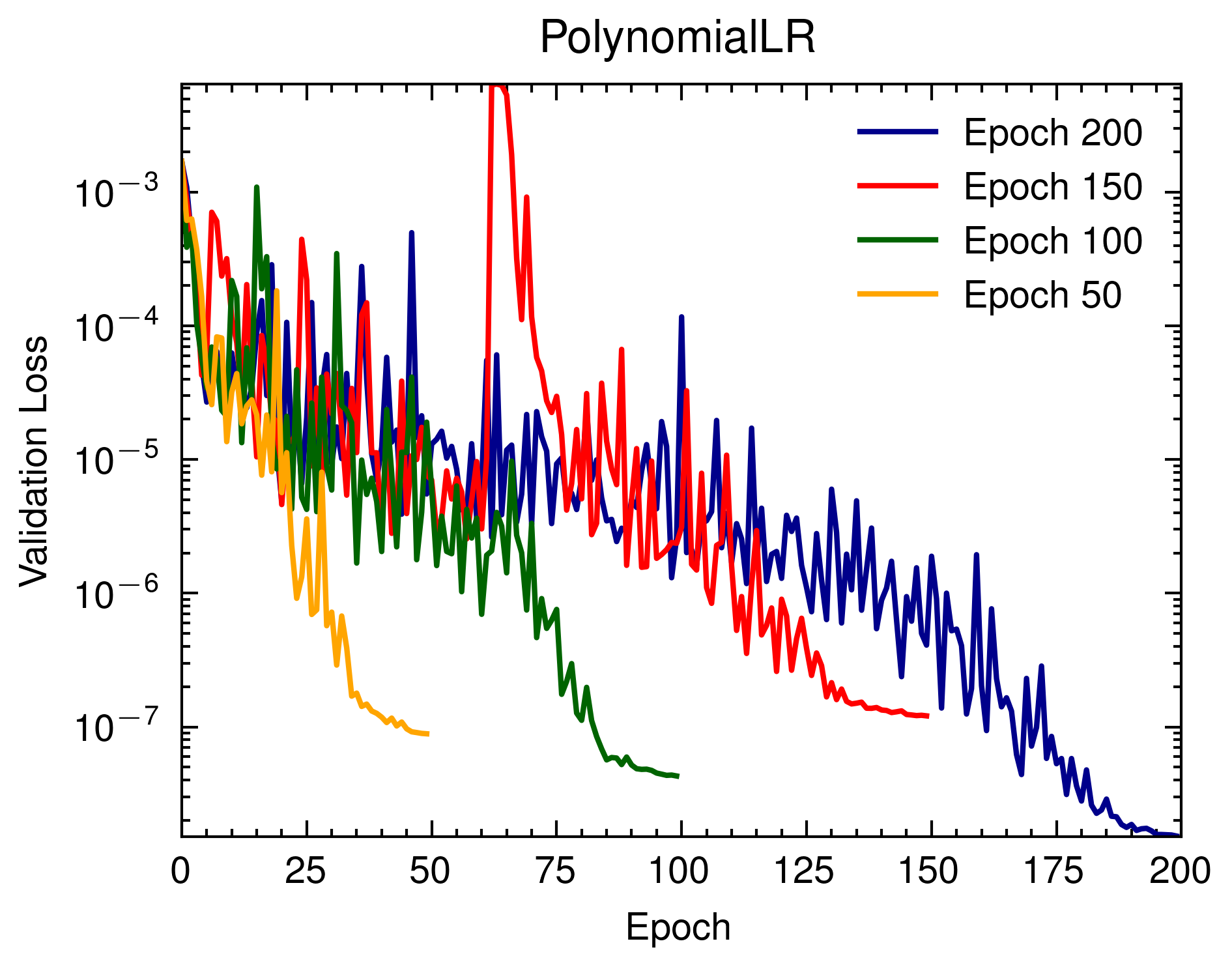}
}\\
\subfloat[]{
    \includegraphics[width=.47\textwidth]{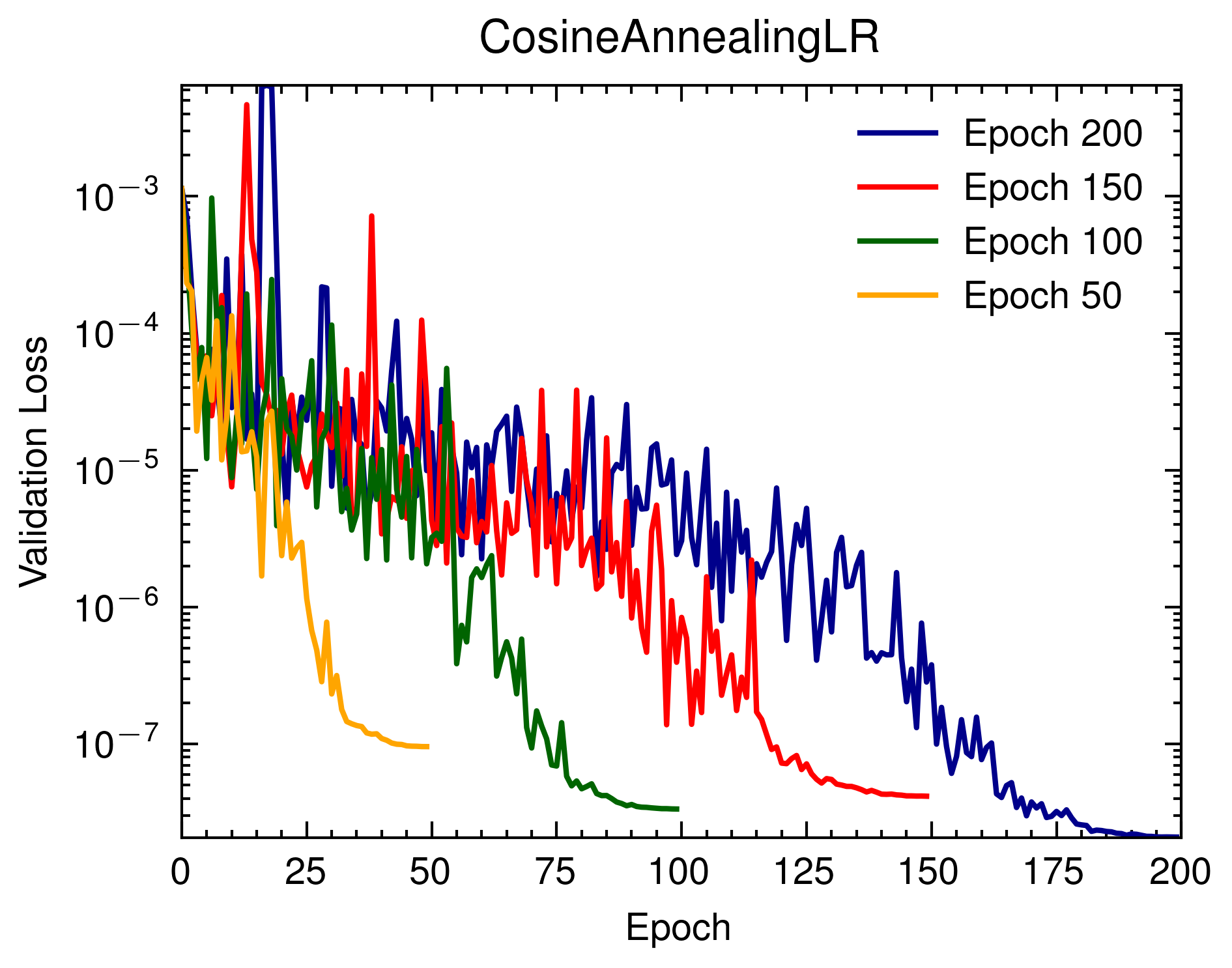}
}
\subfloat[]{
    \includegraphics[width=.47\textwidth]{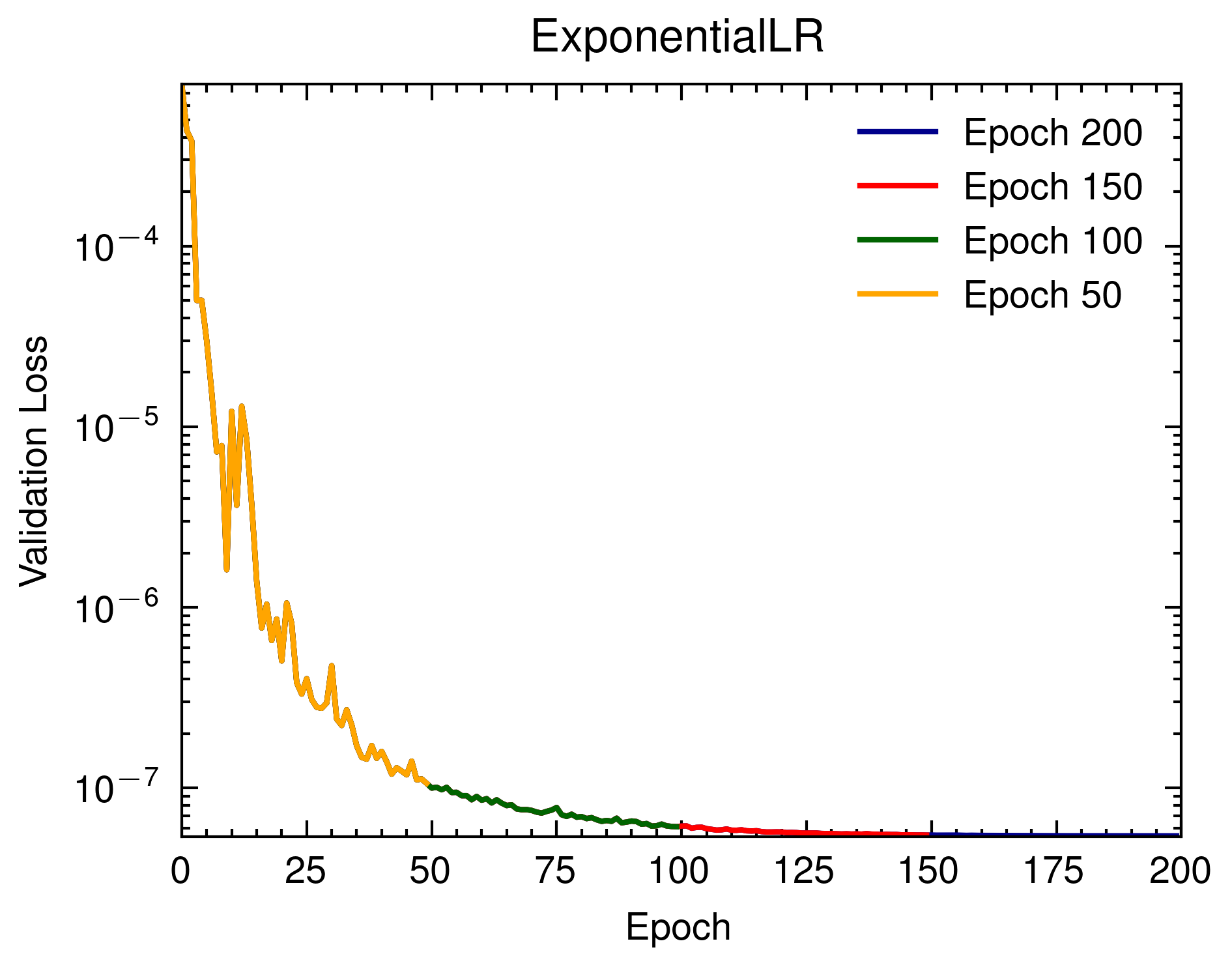}
}\\
\subfloat[]{
    \includegraphics[width=.47\textwidth]{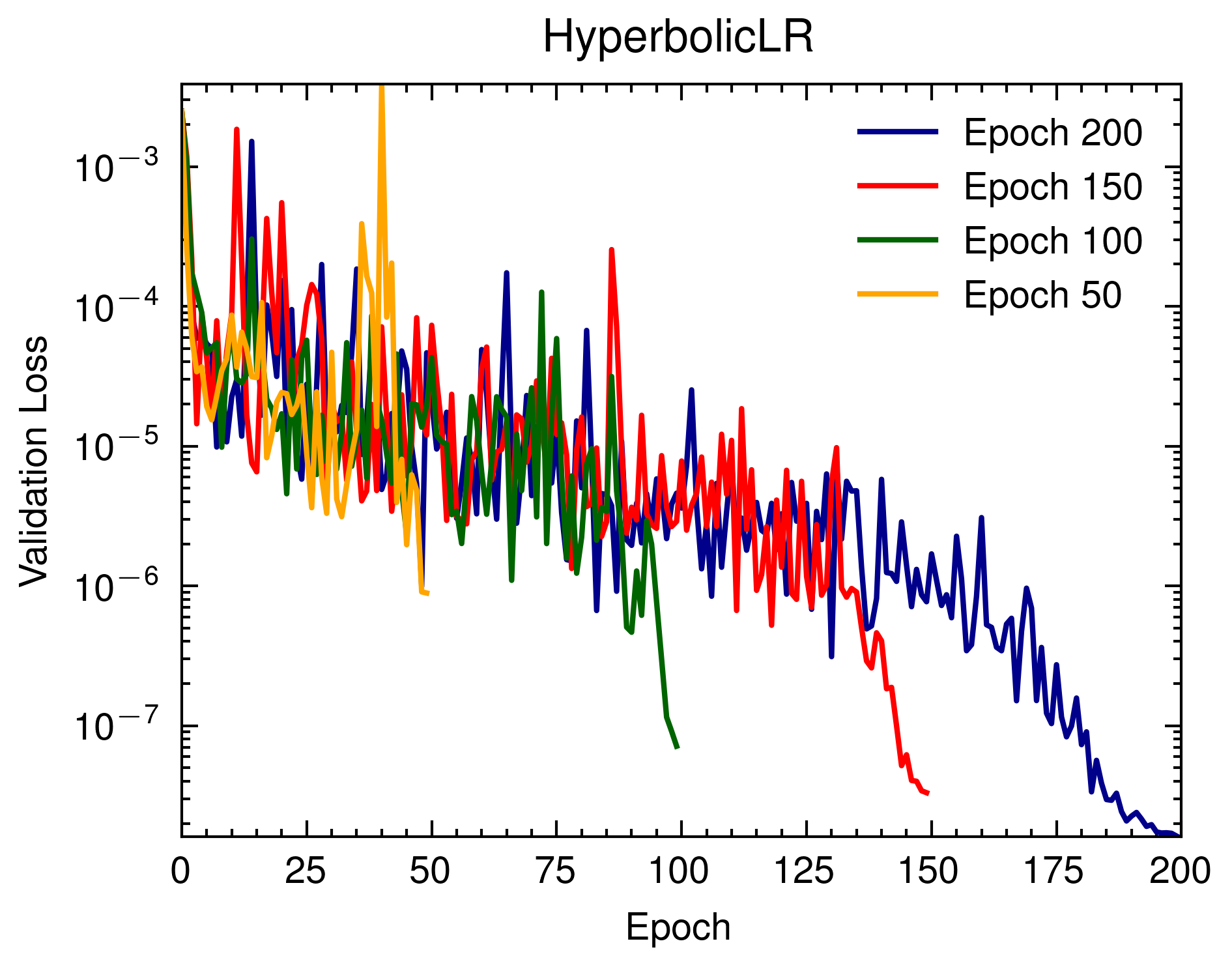}
}
\subfloat[]{
    \includegraphics[width=.47\textwidth]{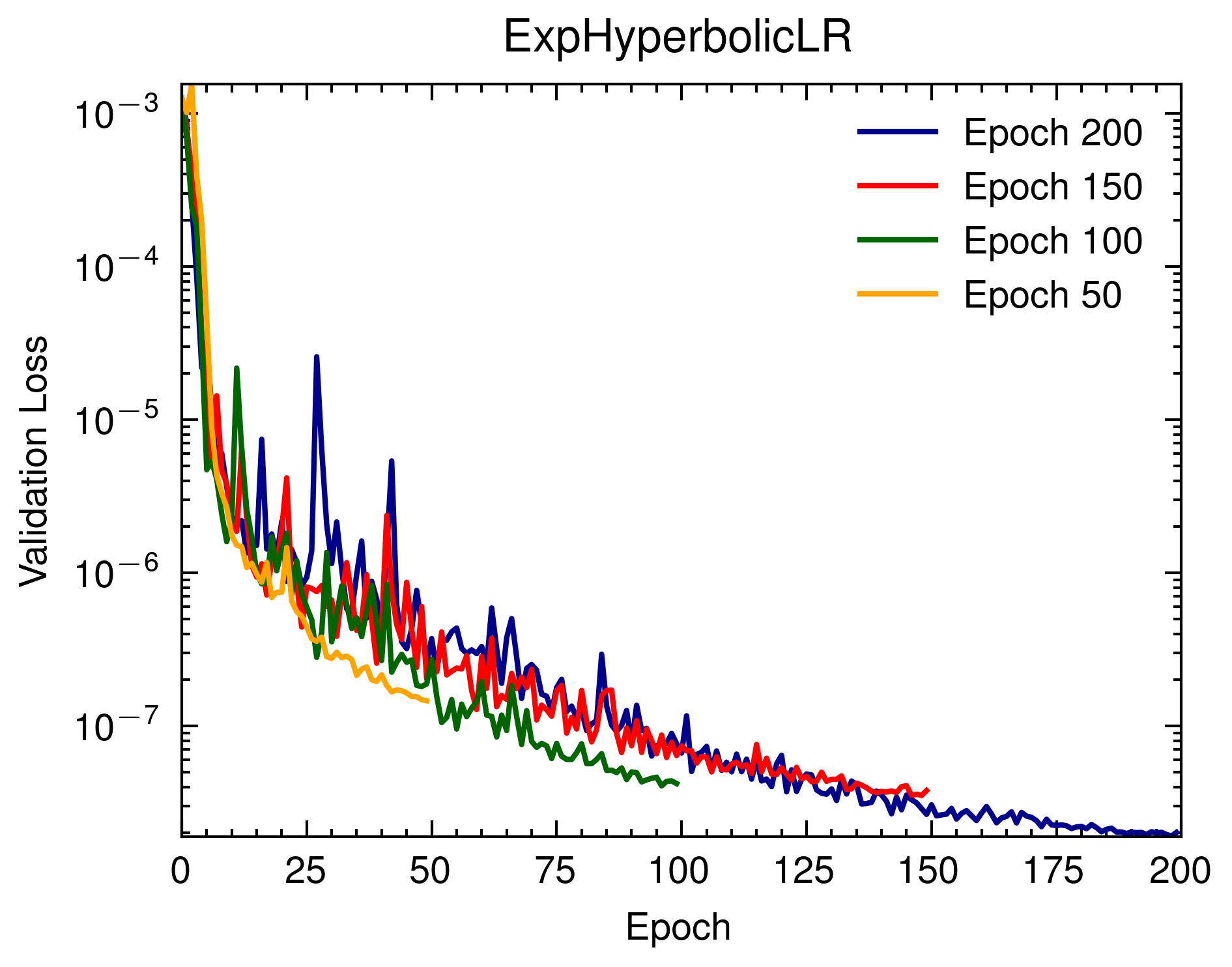}
}
\caption{Learning curves of validation loss for LSTM-Seq2Seq on custom oscillation data.}
\label{fig:learning_curves_OSC_LSTM}
\end{center}
\vskip -0.1in
\end{figure}

\begin{figure}[h]
\captionsetup[subfloat]{labelformat=empty}
\vskip 0.1in
\begin{center}
\subfloat[]{
    \includegraphics[width=.47\textwidth]{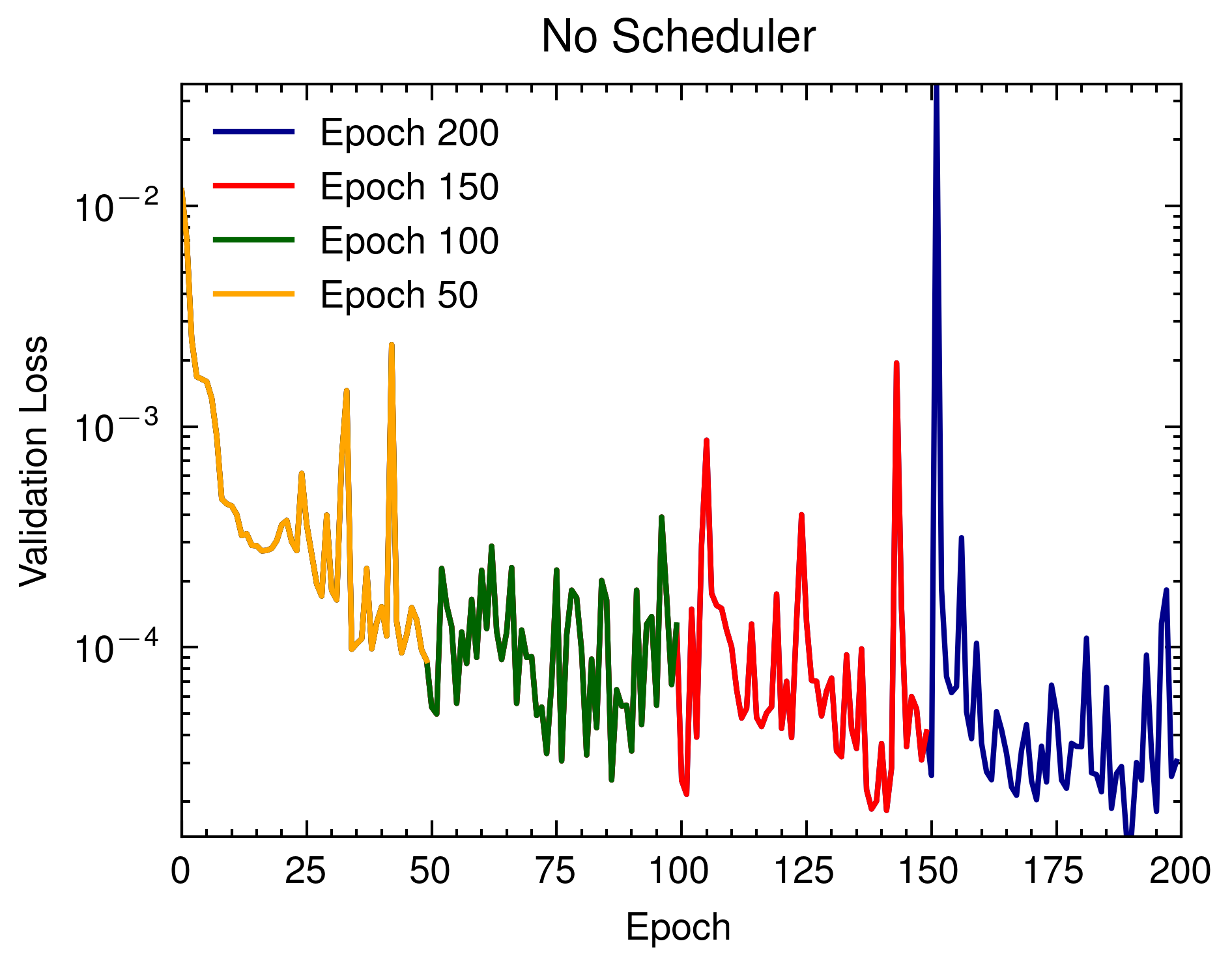}
}
\subfloat[]{
    \includegraphics[width=.47\textwidth]{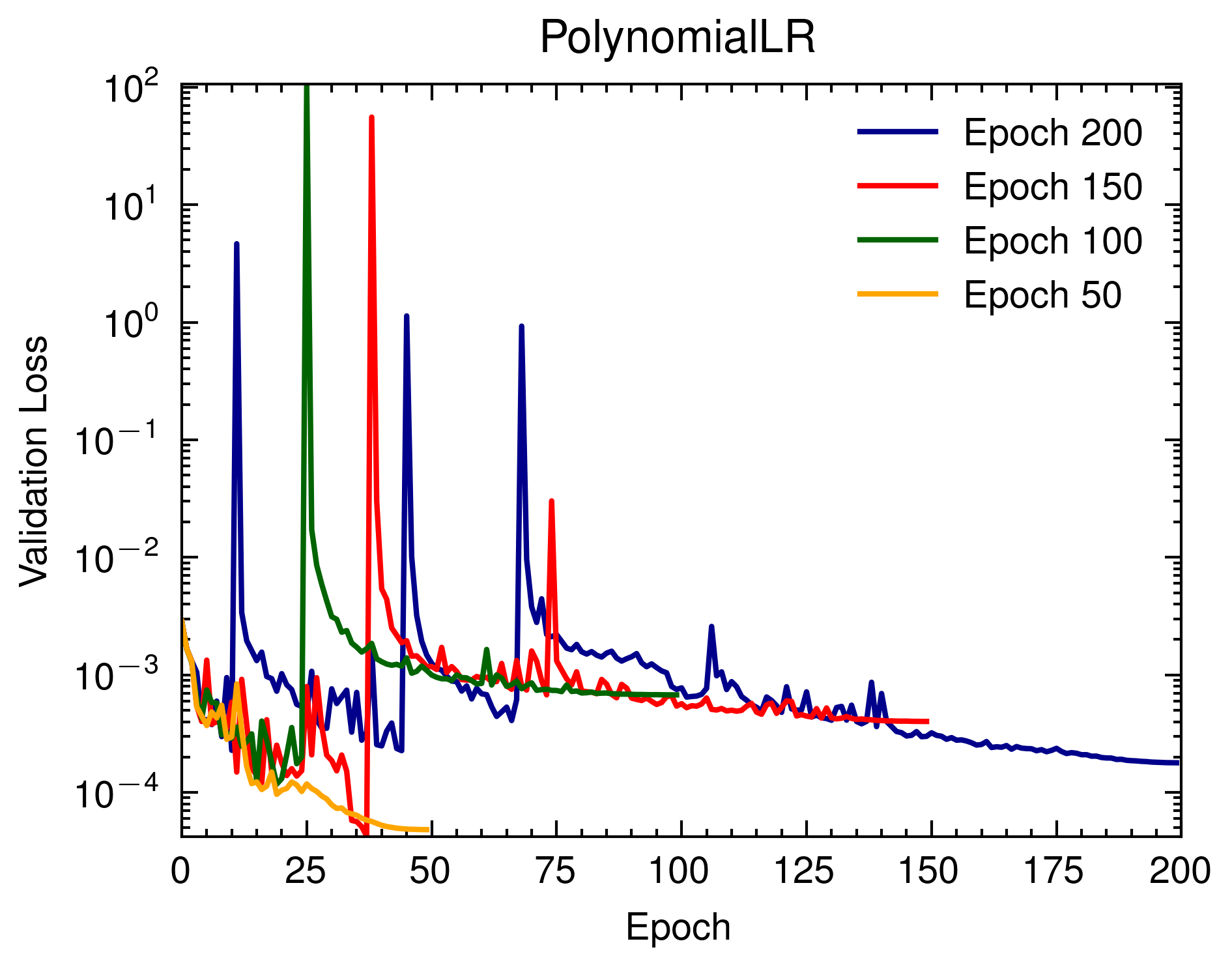}
}\\
\subfloat[]{
    \includegraphics[width=.47\textwidth]{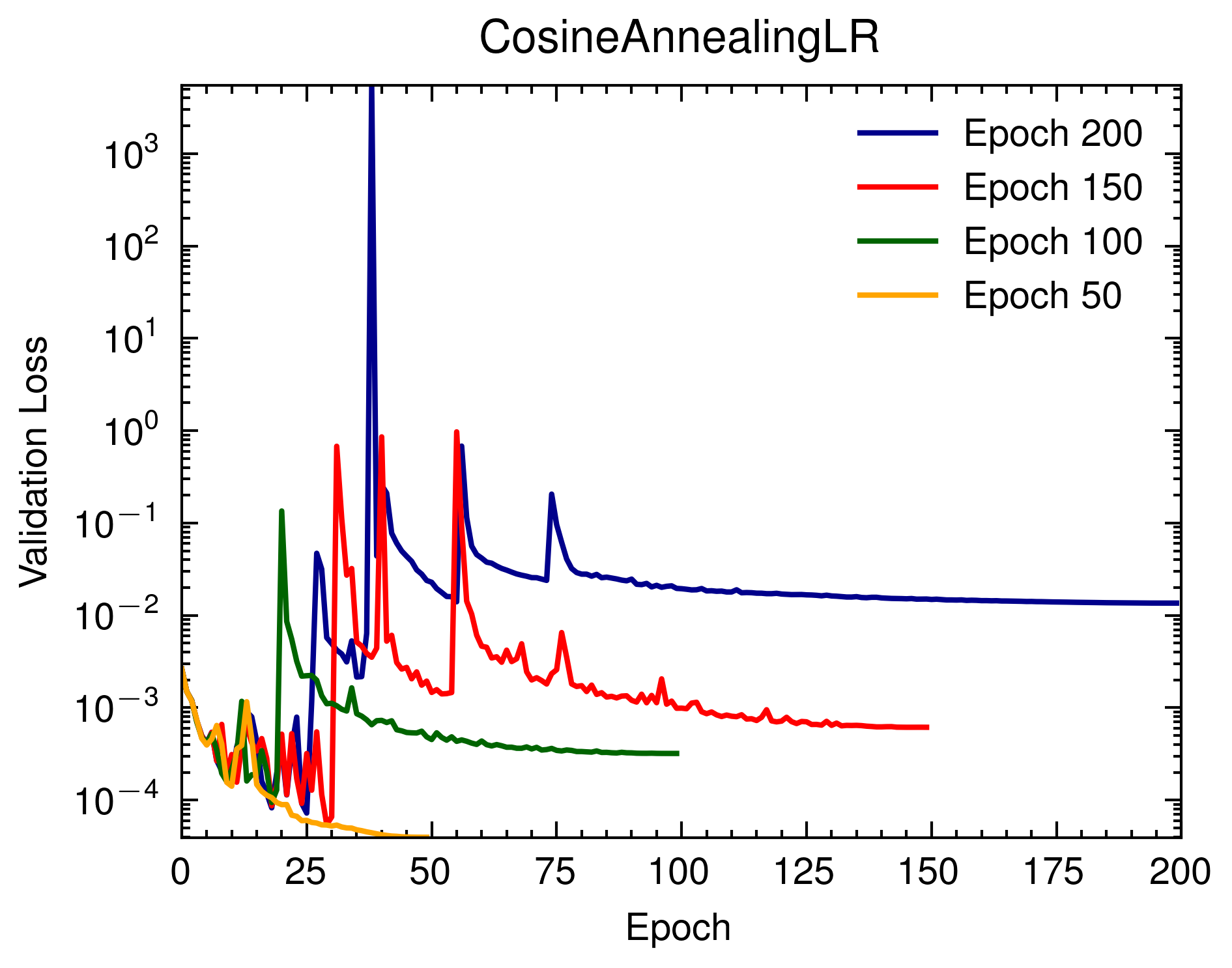}
}
\subfloat[]{
    \includegraphics[width=.47\textwidth]{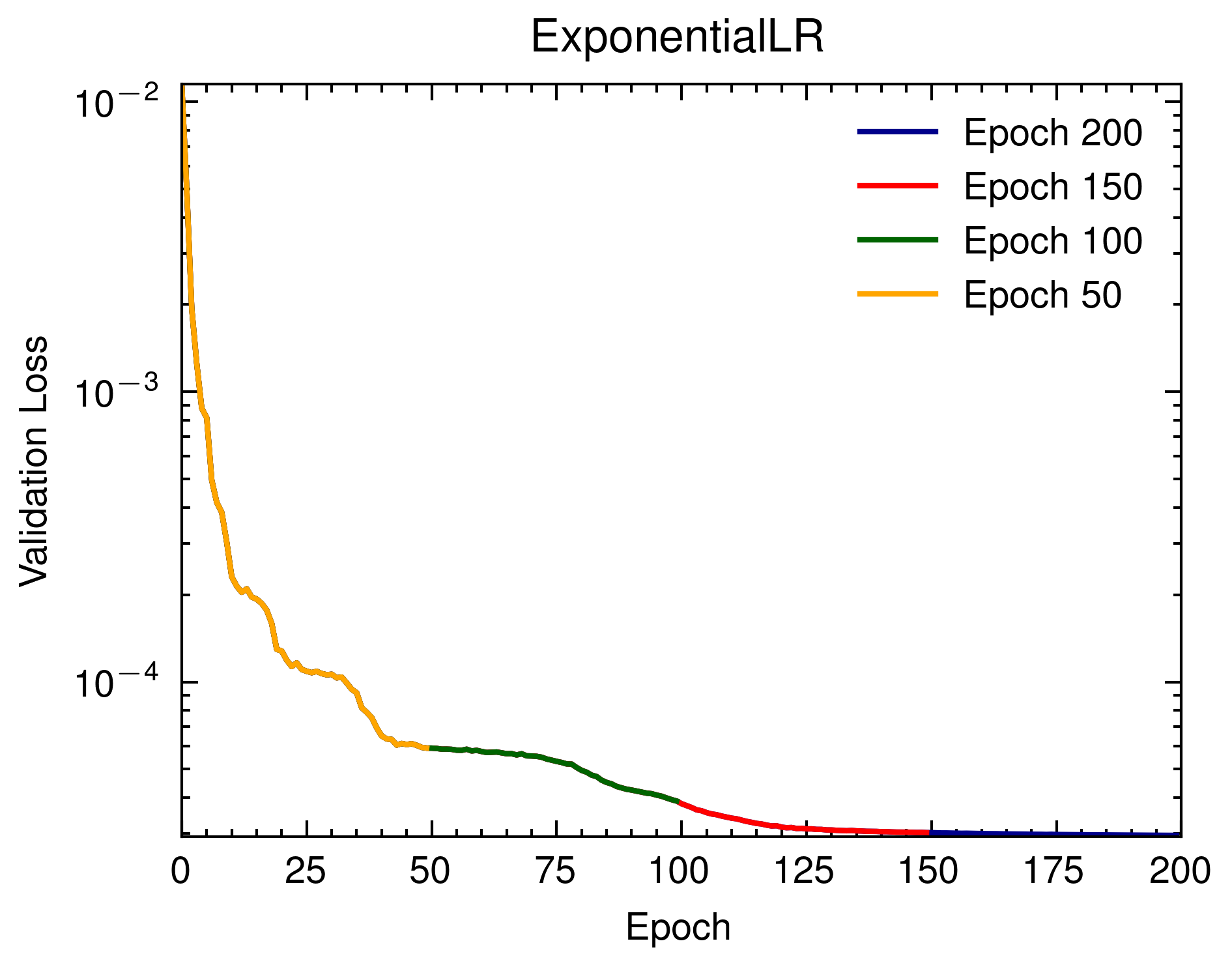}
}\\
\subfloat[]{
    \includegraphics[width=.47\textwidth]{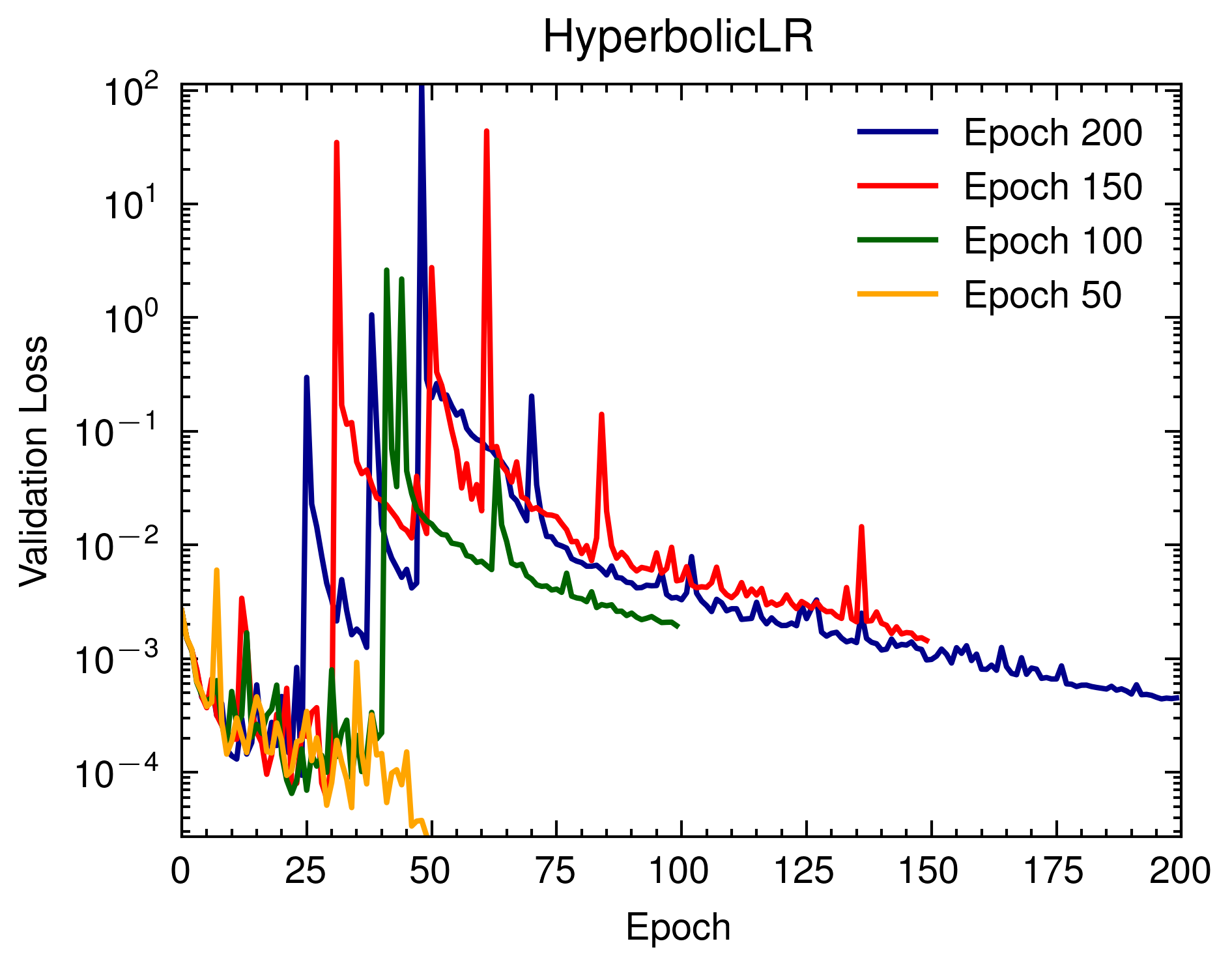}
}
\subfloat[]{
    \includegraphics[width=.47\textwidth]{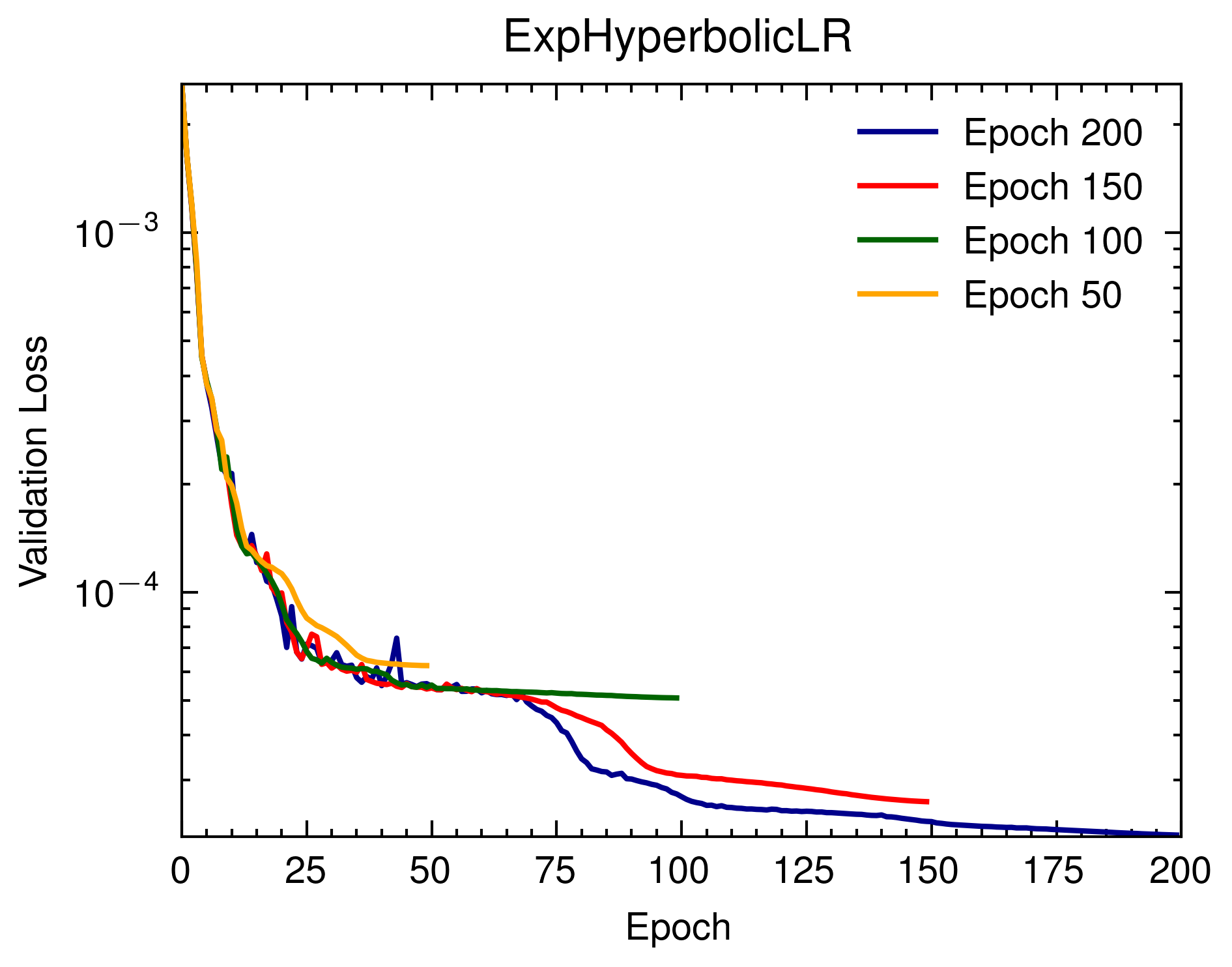}
}
\caption{Learning curves of validation loss for DeepONet on custom integral data.}
\label{fig:learning_curves_Integral_MLP}
\end{center}
\vskip -0.1in
\end{figure}

\begin{figure}[h]
\captionsetup[subfloat]{labelformat=empty}
\vskip 0.1in
\begin{center}
\subfloat[]{
    \includegraphics[width=.47\textwidth]{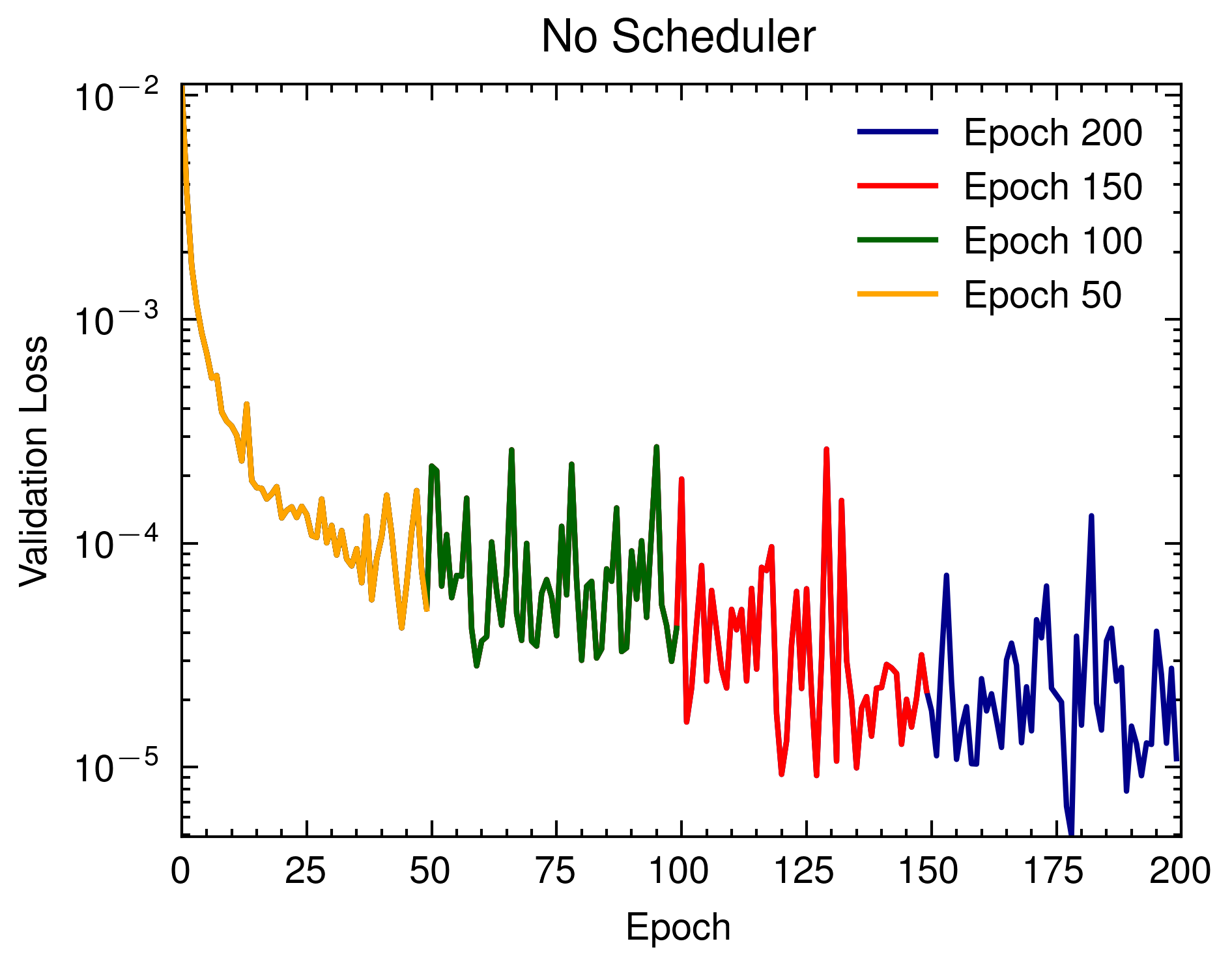}
}
\subfloat[]{
    \includegraphics[width=.47\textwidth]{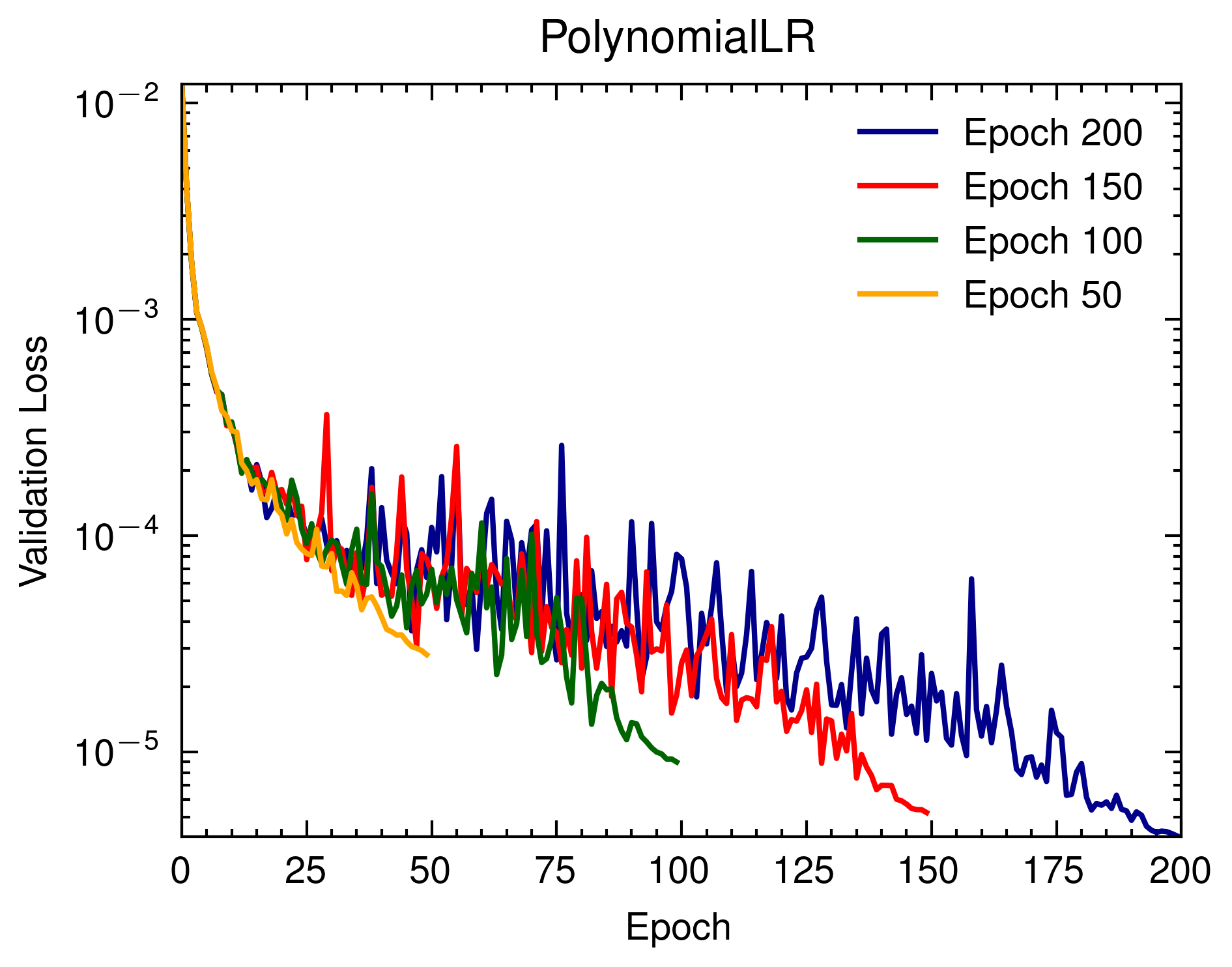}
}\\
\subfloat[]{
    \includegraphics[width=.47\textwidth]{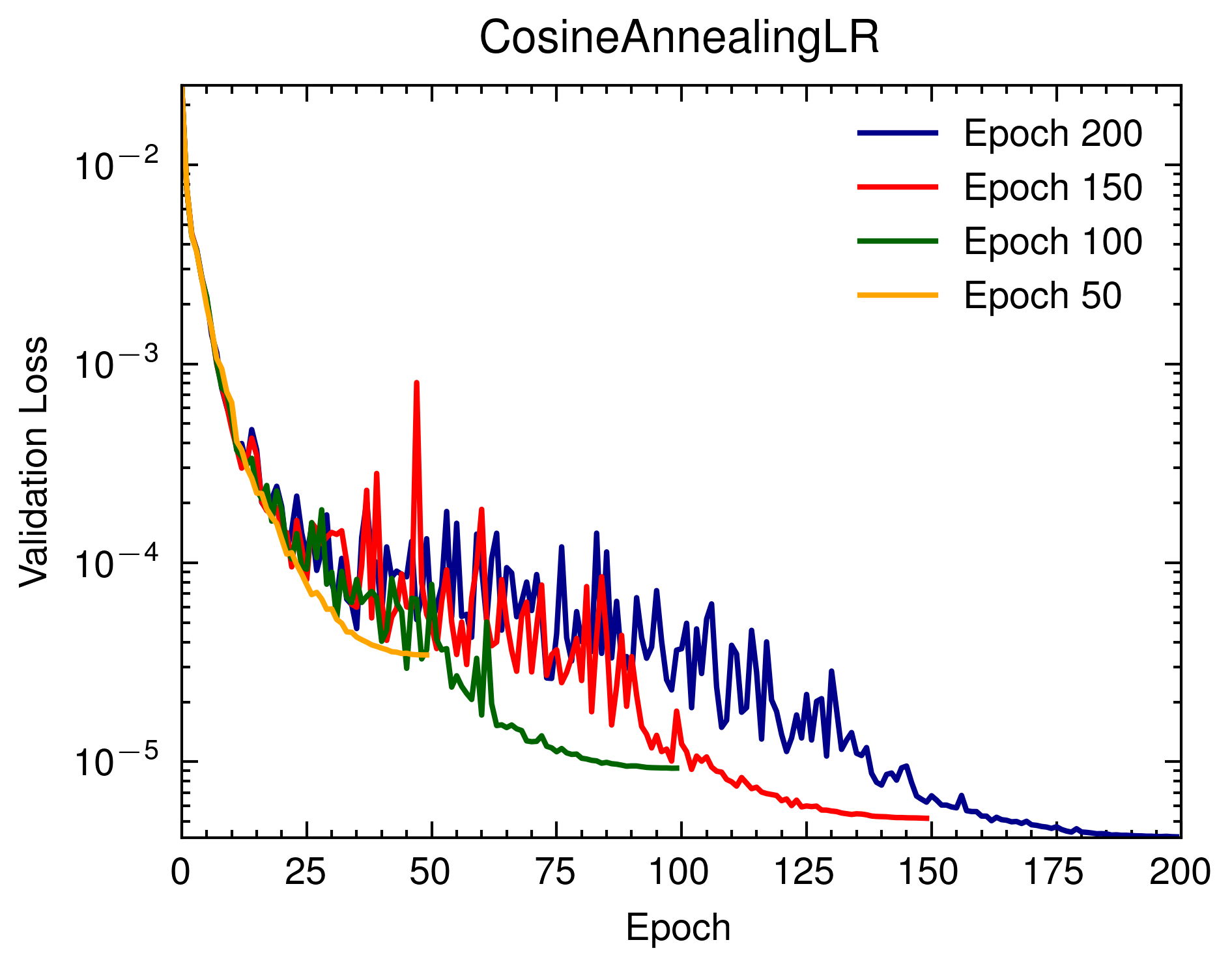}
}
\subfloat[]{
    \includegraphics[width=.47\textwidth]{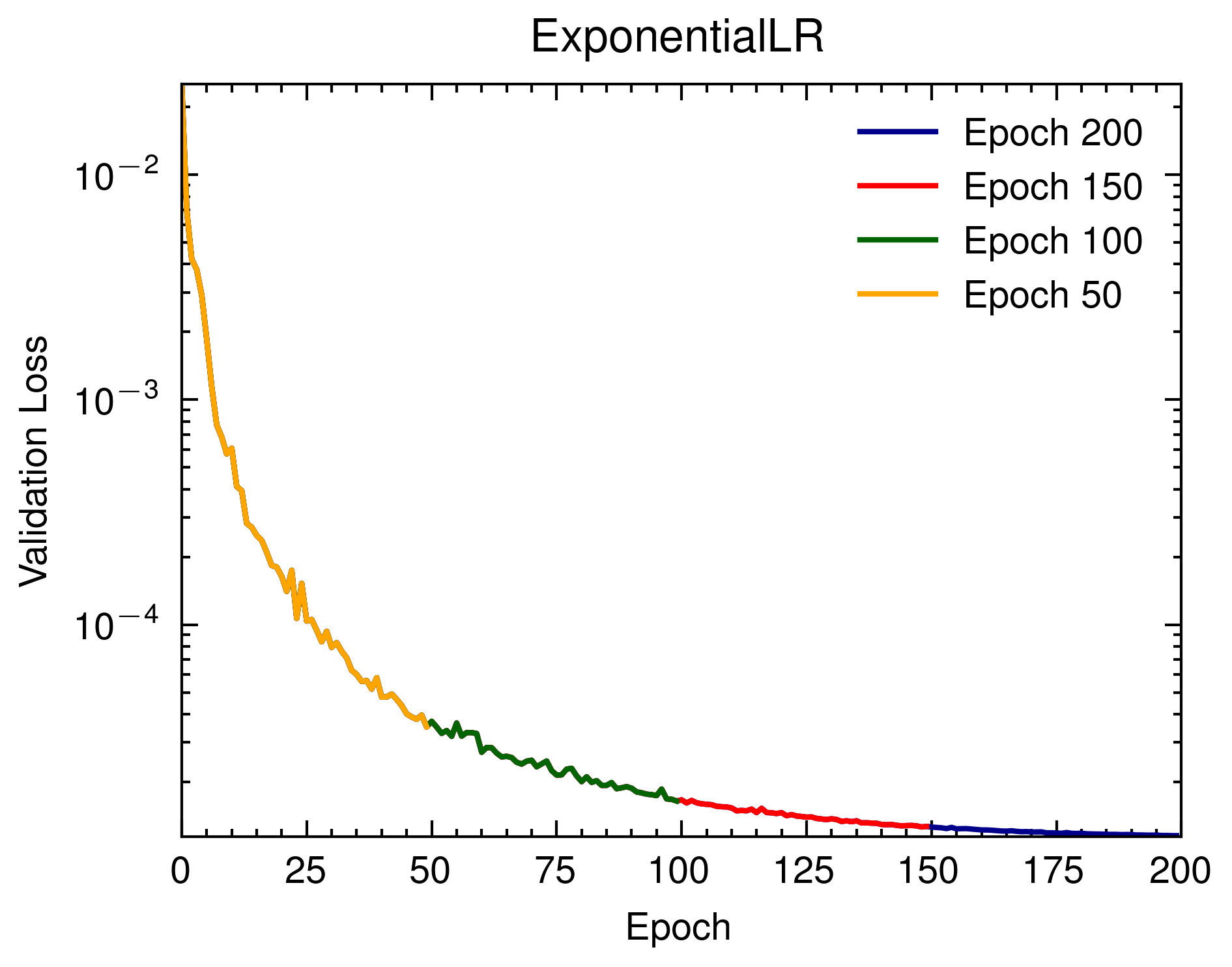}
}\\
\subfloat[]{
    \includegraphics[width=.47\textwidth]{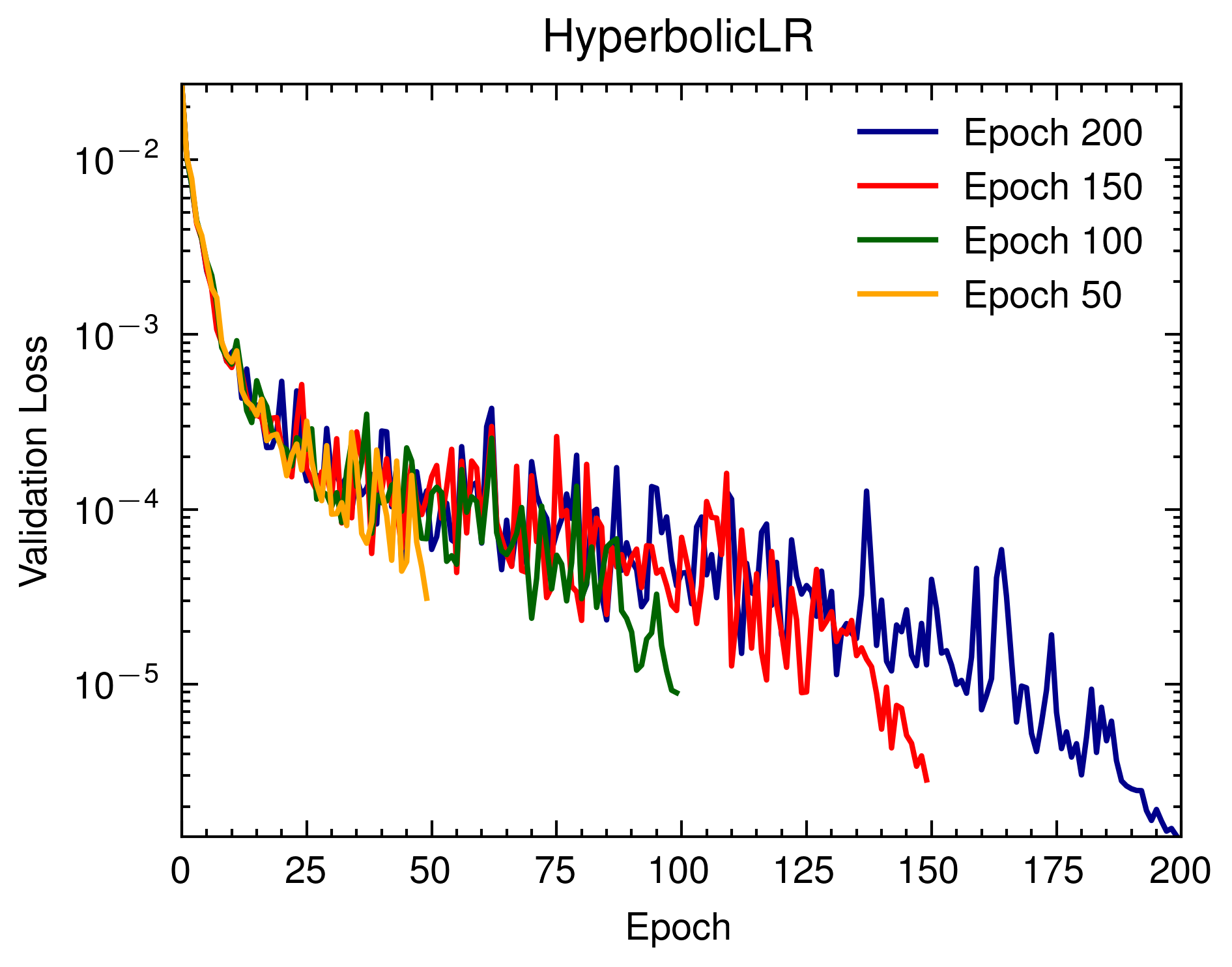}
}
\subfloat[]{
    \includegraphics[width=.47\textwidth]{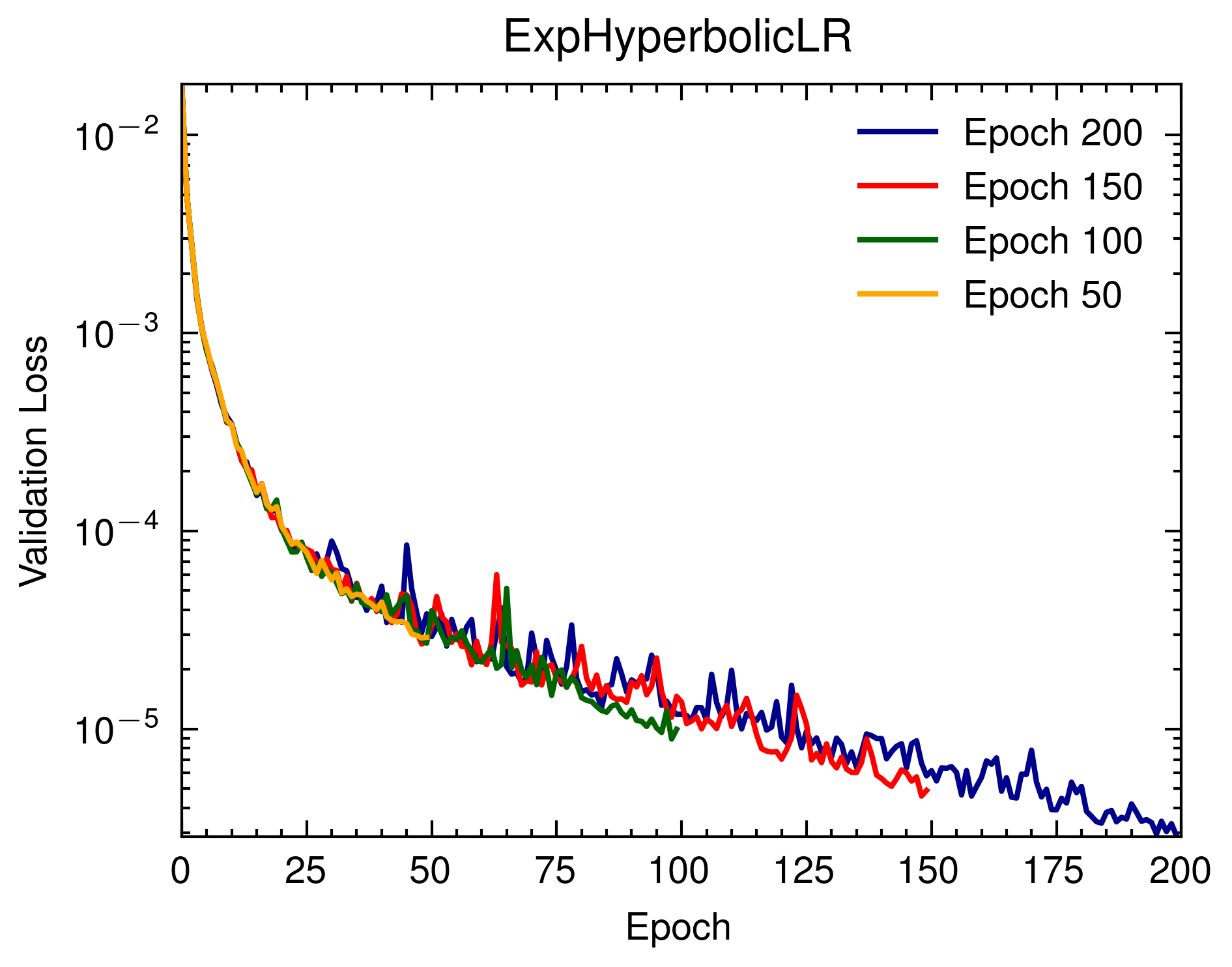}
}
\caption{Learning curves of validation loss for TraONet on custom integral data.}
\label{fig:learning_curves_Integral_TF}
\end{center}
\vskip -0.1in
\end{figure}

To provide a comprehensive view of the training dynamics across different schedulers and tasks, we present a series of learning curves in Figures \ref{fig:learning_curves_CIFAR10_CNN} through \ref{fig:learning_curves_Integral_TF}.
These figures illustrate the validation loss (and accuracy for CIFAR-10) trajectories for each model-scheduler combination over the course of 200 epochs.

Figure \ref{fig:learning_curves_CIFAR10_CNN} and Figure \ref{fig:learning_curves_CIFAR10_CNN_ACC} depict the validation loss and accuracy curves, respectively, for the SimpleCNN model on the CIFAR-10 dataset.
Figures \ref{fig:learning_curves_OSC_LSTM}, \ref{fig:learning_curves_Integral_MLP}, and \ref{fig:learning_curves_Integral_TF} present the validation loss curves for the LSTM Seq2Seq model on the oscillation dataset, and the DeepONet and TraONet models on the integral dataset, respectively.

These visualizations offer insights into the convergence behavior, stability, and relative performance of each scheduler across different tasks and model architectures.
They complement the quantitative analysis presented earlier, providing a qualitative perspective on the learning dynamics throughout the training process.

\end{document}